\documentclass[10pt,twocolumn,letterpaper]{article}

\usepackage[accsupp]{axessibility}  
\usepackage{cvpr}              

\usepackage{graphicx}
\usepackage{amsmath}
\usepackage{amssymb}
\usepackage{booktabs}

\usepackage{paralist}
\usepackage{multirow}
\usepackage{bbding}

\usepackage[pagebackref,breaklinks,colorlinks]{hyperref}

\newcommand\extrafootertext[1]{%
    \bgroup
    \renewcommand\thefootnote{\fnsymbol{footnote}}%
    \renewcommand\thempfootnote{\fnsymbol{mpfootnote}}%
    \footnotetext[0]{#1}%
    \egroup
}

\def\cvprPaperID{4661} 
\def\confName{CVPR}
\def\confYear{2023}

\begin{document}

\title{PREIM3D: 3D Consistent Precise Image Attribute Editing from a Single Image}
\author{Jianhui Li\textsuperscript{\rm 1,2}, Jianmin Li\textsuperscript{\rm 1 $\ast$} , Haoji Zhang\textsuperscript{\rm 1}, Shilong Liu\textsuperscript{\rm 1}, Zhengyi Wang\textsuperscript{\rm 1},  \\
Zihao Xiao\textsuperscript{\rm 3}, Kaiwen Zheng\textsuperscript{\rm 1}, Jun Zhu\textsuperscript{\rm 1 $\ast$}. \vspace{0.2cm} \\
\textsuperscript{\rm 1} Department of Computer Science and Technology, Institute for AI, BNRist, Tsinghua University \\
\textsuperscript{\rm 2} State Key Laboratory of Astronautic dynamics, Xi’an Satellite Control Center \quad  \quad
\textsuperscript{\rm 3} RealAI \vspace{-0.3cm}     
}

\twocolumn[{%
\renewcommand\twocolumn[1][]{#1}%
\maketitle
\begin{center}
    \centering\captionsetup{type=figure}
    \includegraphics[width=\textwidth]{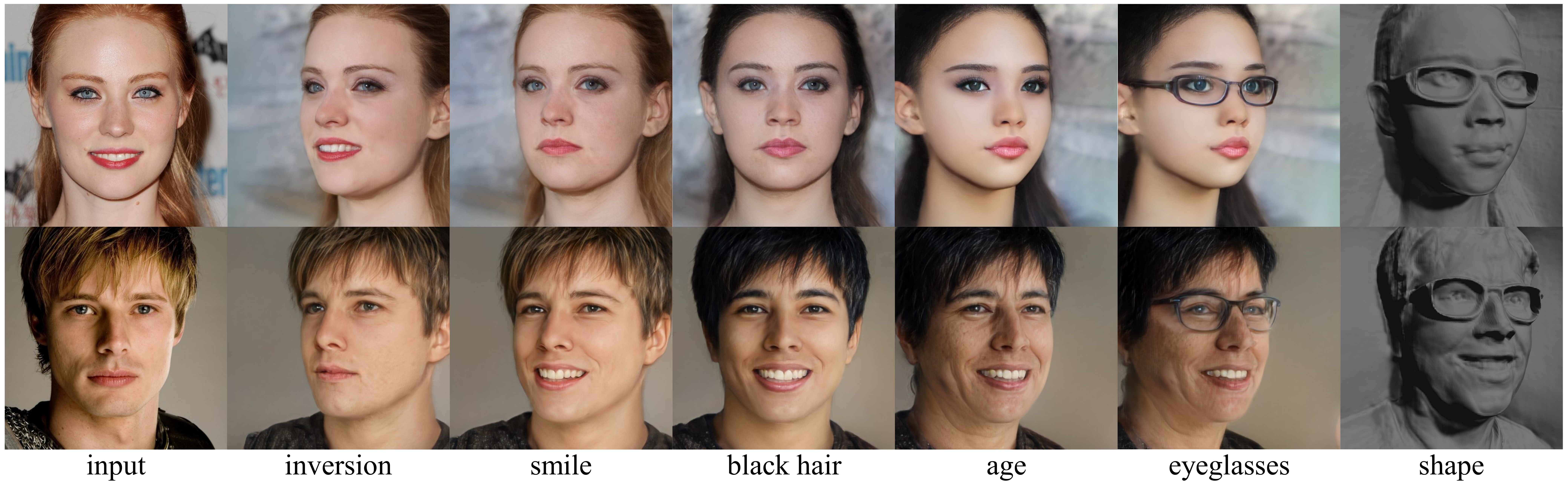}
    \vspace{-0.6cm}
    \captionof{figure}{3D consistent precise inversion and editing. Our method enables reconstructing texture and geometry from a single real image and allows one to perform a list of attributes editing sequentially. The yaw angles of the second to sixth columns are $[-30^{\circ}, -20^{\circ}, 0^{\circ}, 20^{\circ}, 30^{\circ}]$. The last column is the shape of the sixth column.}
    \label{fig:head}
\end{center}%
}]

\extrafootertext{
{{{\textsuperscript{*} Corresponding authors.} }}
}

\begin{abstract}
    We study the 3D-aware image attribute editing problem in this paper, which has wide applications in practice. 
    Recent methods solved the problem by training a shared encoder to map images into a 3D generator's latent space or by per-image latent code optimization and then edited images in the latent space. 
    Despite their promising results near the input view, they still suffer from the 3D inconsistency of produced images at large camera poses and imprecise image attribute editing, like affecting unspecified attributes during editing. 
    For more efficient image inversion, we train a shared encoder for all images. To alleviate 3D inconsistency at large camera poses, we propose two novel methods, an alternating training scheme and a multi-view identity loss, to maintain 3D consistency and subject identity. 
    As for imprecise image editing, we attribute the problem to the gap between the latent space of real images and that of generated images. We compare the latent space and inversion manifold of GAN models and demonstrate that editing in the inversion manifold can achieve better results in both quantitative and qualitative evaluations.
    Extensive experiments show that our method produces more 3D consistent images and achieves more precise image editing than previous work.
    Source code and pretrained models can be found on our project page: \url{https://mybabyyh.github.io/Preim3D/}.
\end{abstract}


\section{Introduction}
\label{sec:intro}
Benefiting from the well-disentangled latent space of Generative Adversarial Networks (GANs)~\cite{goodfellow2020generative}, many works study GAN inversion~\cite{abdal2019image2stylegan,abdal2020image2stylegan++,tov2021designing,richardson2021encoding,zhu2020domain,wang2022high,DBLP:conf/cvpr/Dinh0NH22}  as well as real image editing in the latent space~\cite{harkonen2020ganspace,shen2020interfacegan,shen2021closed,hu2022style,ling2021editgan}.
With the popularity of Neural Radiance Fields (NeRF)~\cite{mildenhall2021nerf}, some works start to incorporate it into GAN frameworks for unconditional 3D-aware image generation~\cite{schwarz2020graf,niemeyer2021giraffe,nguyen2019hologan,chan2021pi,chan2022efficient,gu2021stylenerf,or2022stylesdf}. 
In particular, EG3D~\cite{chan2022efficient}, the state-of-the-art 3D GAN, is able to generate high-resolution multi-view-consistent images and high-quality geometry conditioned on gaussian noise and camera pose.
Similar to 2D GANs, 3D GANs also have a well semantically disentangled latent space~\cite{chan2022efficient,lin20223d,gu2021stylenerf,sun2022ide}, which enables realistic yet challenging 3D-aware image editing.

Achieving 3D-aware image editing is much more challenging because it not only has to be consistent with the input image at the input camera pose but also needs to produce 3D consistent novel views. 
Recently, 3D-Inv~\cite{lin20223d} uses pivotal tuning inversion (PTI) ~\cite{roich2022pivotal}, first finding out a pivotal latent code and then finetuning the generator with the fixed pivotal latent code, to obtain the latent code and edit the image attributes in the latent space.
IDE-3D~\cite{sun2022ide} proposes a hybrid 3D GAN inversion approach combining texture and semantic encoders and PTI technique, accelerating the optimization process by the encoded initial latent code. 
Pixel2NeRF ~\cite{cai2022pix2nerf} is the first to achieve 3D inversion by training an encoder mapping a real image to the latent space $\mathcal{Z}$ of $\pi$-GAN~\cite{chan2021pi}.
However, these methods still do not solve the problem of 3D consistency at large camera poses and precise image attribute editing. As shown in Fig. \ref{fig:quality_compare}, some inverted images meet head distortion at large camera poses, or some unspecific attributes of edited images are modified.  

In this paper, we propose a pipeline that enables \textbf{PR}ecise \textbf{E}diting in the \textbf{I}nversion \textbf{M}anifold with \textbf{3D} consistency efficiently, termed \textbf{PREIM3D}. There are three goals to achieve for our framework, (i) image editing \textit{efficiently}, (ii) \textit{precise inversion}, which aims to maintain realism and 3D consistency of multiple views, and (iii) \textit{precise editing}, which is to edit the desired attribute while keeping the other attributes unchanged.
3D-Inv and IDE-3D optimized a latent code for each image, which is not suitable for interactive applications. Following Pixel2NeRF, we train a shared encoder for all images for efficiency.

To address \textit{precise inversion}, we introduce a 3D consistent encoder to map a real image into the latent space $\mathcal{W}^+$ of EG3D, and it can infer a latent code with a single forward pass.
We first design a training scheme with alternating in-domain images (i.e., the generated images) and out-domain images (i.e., the real images) to help the encoder maintain the 3D consistency of the generator.
We optimize the encoder to reconstruct the input images in the out-domain image round.
In the in-domain image round, we additionally optimize the encoder to reconstruct the ground latent code, which will encourage the distribution of the inverted latent code closer to the distribution of the original latent code of the generator.
Second, to preserve the subject's identity, we propose a multi-view identity loss calculated between the input image and novel views randomly sampled in the surrounding of the input camera pose.

Though many works tried to improve the editing precision by modifying latent codes in $\mathcal{Z}$ space~\cite{shen2020interfacegan}, $\mathcal{W}$ space~\cite{harkonen2020ganspace,jahanian2019steerability,tewari2020stylerig}, $\mathcal{W}^+$ space~\cite{abdal2019image2stylegan,abdal2019image2stylegan,zhu2020domain}, and $\mathcal{S}$ space~\cite{wu2021stylespace}, they all still suffer from a gap between real image editing and generated image editing because of using the editing directions found in the original generative latent space to edit the real images.
To bridge this gap, we propose a real image editing subspace, which we refer to \textit{inversion manifold}.
We compare the inversion manifold and the original latent space and find the distortion between the attribute editing directions.
We show that the editing direction found in the inversion manifold can control the attributes of the real images more precisely.
To our knowledge, we are the first to perform latent code manipulation in the inversion manifold.
Our methodology is orthogonal to some existing editing methods and can improve the performance of manipulation in qualitative and quantitative results when integrated with them.
Figure \ref{fig:head} shows the inversion and editing results produced by our method.
Given a single real image, we achieve 3D reconstruction and precise multi-view attribute editing. 

The contributions of our work can be summarized as follows:
\begin{itemize}
\setlength{\itemsep}{3pt}
\setlength{\parsep}{3pt}
\setlength{\parskip}{3pt}
    \item We present an efficient image attribute editing method by training an image-shared encoder for 3D-aware generated models in this paper. To keep 3D consistency at large camera poses, we propose two novel methods, an alternating training scheme and a multi-view identity loss, to maintain 3D consistency and subject identity. 
    \item We compare the latent space and inversion manifold of GAN models, and demonstrate that editing in the inversion manifold can achieve better results in both quantitative and qualitative evaluations. The proposed editing space helps to close the gap between real image editing and generated image editing.
    \item We conduct extensive experiments, including both quantitative and qualitative, on several datasets to show the effectiveness of our methods. 
\end{itemize}

\section{Related Work}
\label{sec:related}
\subsection{NeRF-based GANs}
NeRF models the underlying 3D scene through a continuous 5D function $F_\Theta$ that maps point ($x,y,z$) and viewing direction ($\theta, \phi$) to color and corresponding density, and then uses volume rendering techniques to render multi-view images~\cite{mildenhall2021nerf}.
Although the standard NeRF requires multi-view images and trains the network for every single scene, several works ~\cite{schwarz2020graf,niemeyer2021giraffe,nguyen2019hologan,chan2021pi,gu2021stylenerf,chan2022efficient} combine it with the GAN framework to generate multi-view images from unconditional random samples.
Among these, HoloGAN~\cite{nguyen2019hologan} is the first NeRF-based GAN to learn 3D representations from unposed 2D images in an unsupervised manner.
GRAF~\cite{schwarz2020graf} produces high-resolution multi-view images of novel objects from disentangled shape code $z_s$ and appearance code $z_a$.
GIRAFFE~\cite{niemeyer2021giraffe} can handle multi-object 3D scenes and control the synthesis of all objects separately.
StyleNeRF~\cite{gu2021stylenerf} and EG3D~\cite{chan2022efficient} both utilize 2D CNN upsampling after neural rendering feature to achieve high-resolution 3D-aware images.
Especially, EG3D~\cite{chan2022efficient} uses the triplane representation to perform volume rendering, which is much more efficient than fully 3D networks.
Our work is based on EG3D, which has a semantically disentangled latent space comparable to the state-of-the-art 2D generator, StyleGAN.

\begin{table*}
  \centering
  \resizebox{0.7\linewidth}{!}{%
  \begin{tabular}{ccccc}
    \toprule
     & Inversion Type & Efficiently & 3D Consistency at Large Camera Poses &  Superior Editability\\ 
    \midrule
    IDE-3D~\cite{sun2022ide}  & hybrid & \XSolidBrush &\XSolidBrush  &\XSolidBrush  \\
    3D-Inv~\cite{lin20223d}   & optimization-based  &\XSolidBrush  &\XSolidBrush  &\XSolidBrush \\
    Pixel2NeRF~\cite{cai2022pix2nerf}& encoder-based  & \checkmark &\XSolidBrush &\XSolidBrush \\
    PREIM3D (Ours)                   & encoder-based  & \checkmark & \checkmark & \checkmark \\
    \bottomrule
  \end{tabular}}
  \vspace{-0.3cm}
  \caption{An overview of 3D GAN inversion and editing methods.
  3D consistency and editability are evaluated by ID, APD, AA, and AD metrics described in section \ref{sec:experiments}.
  }
  \vspace{-0.3cm}
  \label{tab:method_feature}
\end{table*}

\subsection{NeRF-based GAN Inversion}
With the great success of GAN in the area of image generation, GAN inversion has become a popular research topic in image editing tasks.
Existing GAN inversion methods can be grouped into three categories: optimization-based, encoder-based, and hybrid both.
Optimization methods~\cite{abdal2019image2stylegan,abdal2020image2stylegan++,creswell2018inverting,karras2020analyzing} directly optimize the latent code to minimize the distance between the generated image and the given image.
Instead of optimization for every image, the encoder methods~\cite{tov2021designing,richardson2021encoding,alaluf2021restyle,hu2022style} train a category-specific generic encoder to map the given image to the latent code.
pSp~\cite{richardson2021encoding} extracts features from different pyramid scales encoding the image to style codes in $\mathcal{W}^+$ space.
Further, e4e~\cite{tov2021designing} analyzes the trade-off between distortion and editability.
Based on e4e, HFGI~\cite{wang2022high}  injects distortion residual features into the generator to improve the fidelity.
Hybrid approach is a combination of encoder and optimization, leveraging the encoder to produce the initial latent code for optimization.
The optimization-based and hybrid methods tend to favor faithful reconstruction over editability~\cite{richardson2021encoding,wang2022high}.

Despite the success of GAN inversion in 2D space, it is at its early age in 3D space.
3D GAN inversion is required to "imagine" the 3D geometry given only one single image, which is much more difficult than 2D GAN inversion.
Recently, 3D-Inv~\cite{lin20223d} adopts PTI~\cite{roich2022pivotal}, a two-stage method, to perform the inversion. 
In the first stage, they optimize for the latent code $w$ reconstructing the face region by freezing the generator.
In the second stage, they freeze the optimized latent code $w^*$, pivotal latent code, and fine-tune the generator by minimizing the LPIPS~\cite{zhang2018unreasonable} loss.
Further,IDE-3D~\cite{sun2022ide} combines encoder and PTI, leveraging texture and semantic encoders to produce the pivotal latent code for acceleration.
Since the PTI-based method requires optimization for each given image, it cost a long inference time and may converge to an arbitrary point in the latent space, which is not conducive to editing.
Moreover, optimization on a single image leads to artifacts in novel views.
In contrast, our method is a pure encoder requiring very little inference time, maintaining the perfect 3D consistency of the pretrained generator by fixing it.

Pixel2NeRF~\cite{cai2022pix2nerf} is the first attempt to train a pure encoder mapping a single real image to the $\mathcal{Z}$ latent space in 3D GAN inversion task, which became a strong baseline.
However, Pixel2NeRF suffers from attribute entanglement, low resolution, and 3D inconsistency.
Different from Pixel2Nerf, we map the input image to $\mathcal{W}^+$ space, which is a more disentangled and editable space~\cite{karras2020analyzing,wu2021stylespace}, and improve the 3D consistency by explicitly encouraging multiple views. Table \ref{tab:method_feature} outlines the differences between our method and previous methods.

\subsection{Latent Space Manipulation}
Numerous works have explored the GAN latent space for performing semantically manipulations in supervised or unsupervised manners~\cite{collins2020editing,shen2020interfacegan,harkonen2020ganspace, abdal2021styleflow, tewari2020stylerig,jahanian2019steerability,DBLP:conf/iclr/WuNSL22}.
The common approach is to seek the editing direction responsible for controlling a given attribute.
GANSpace~\cite{harkonen2020ganspace} adopts principal component analysis (PCA) to find the semantic direction.
Sefa~\cite{shen2021closed} directly decomposes the weight of the pretrained mapping network to discover the latent semantics with a closed-form factorization algorithm.
InterfaceGAN~\cite{shen2020interfacegan} uses the normal vector of the boundary hyperplane of the SVM as the editing direction of the binary attribute.
StyleSpace~\cite{wu2021stylespace} proposes to use style channels to control a specific target attribute in $\mathcal{S}$ space which is defined by the channel-wise style parameters.
To further disentangled the attribute editing, StyleFlow~\cite{abdal2021styleflow} trains a flow network formulating attribute-conditioned exploration as an instance of conditional continuous normalizing flows.
These methods edit images in the original latent space of the generator, but we find that editing images in the inversion manifold achieves better qualitative and quantitative results.

\section{Method}

\begin{figure*}[htb]
  \centering
  \begin{subfigure}{1.0\linewidth}
    \includegraphics[width=1.0\textwidth]{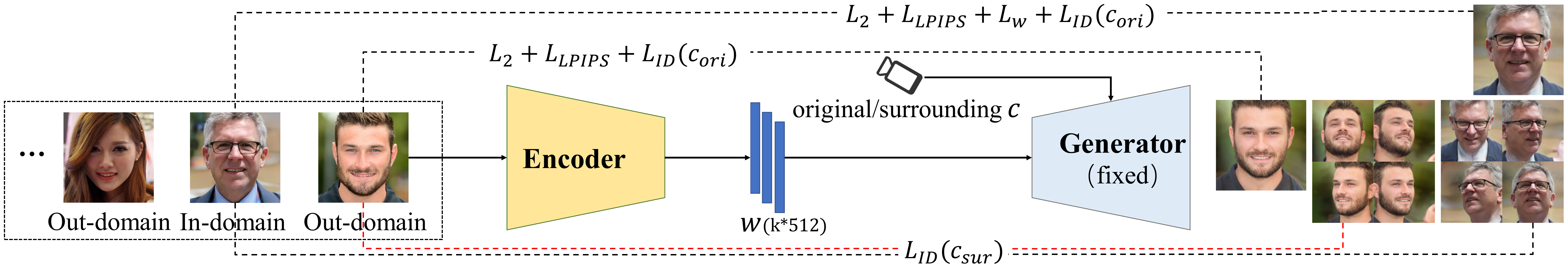}
    \caption{}
    \label{fig:pipeline-a}
  \end{subfigure}
  \begin{subfigure}{0.40\linewidth}
    \includegraphics[width=1.0\textwidth]{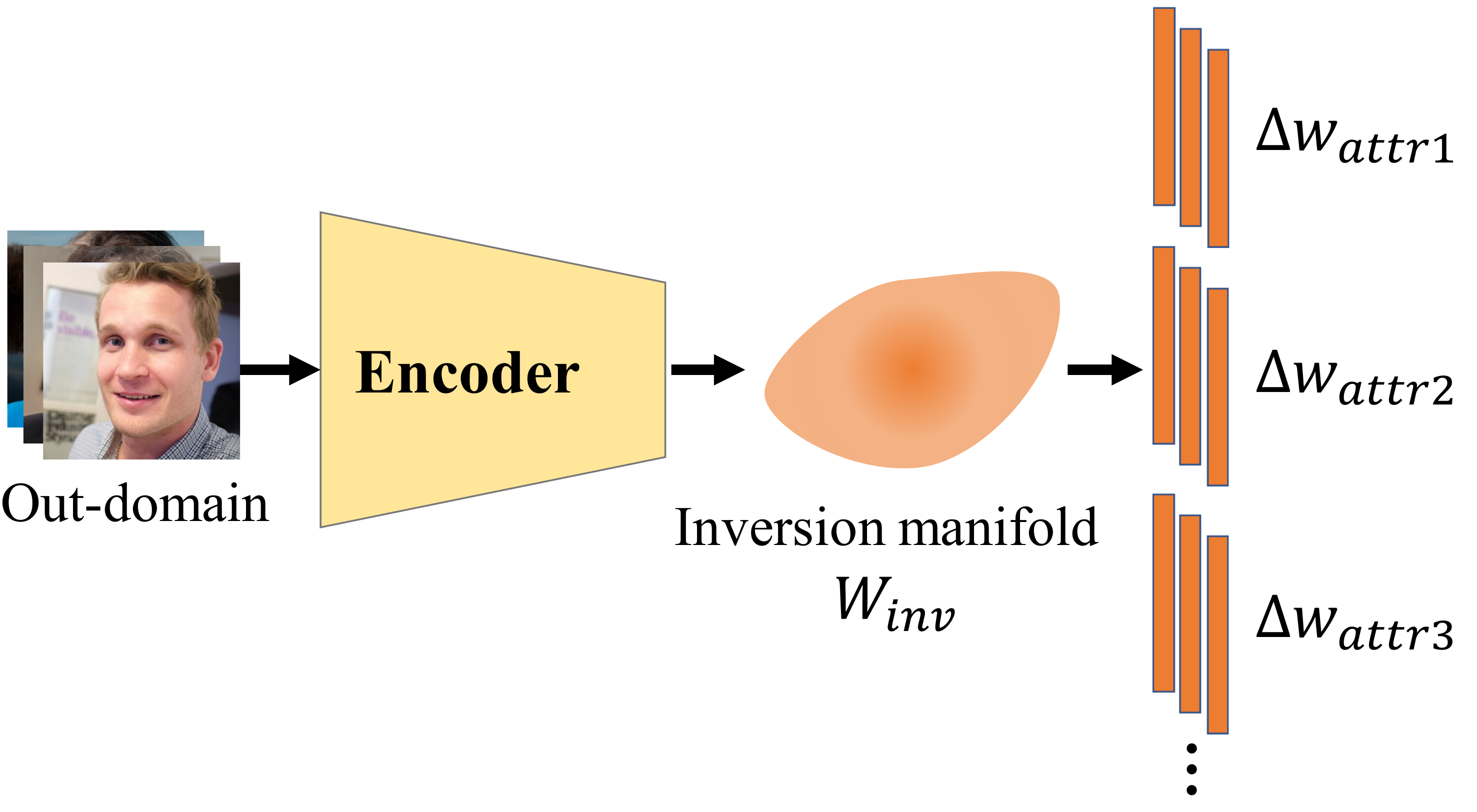}
    \caption{}
    \label{fig:pipeline-b}
  \end{subfigure}
  \hfill
  \begin{subfigure}{0.50\linewidth}
    \includegraphics[width=1.0\textwidth]{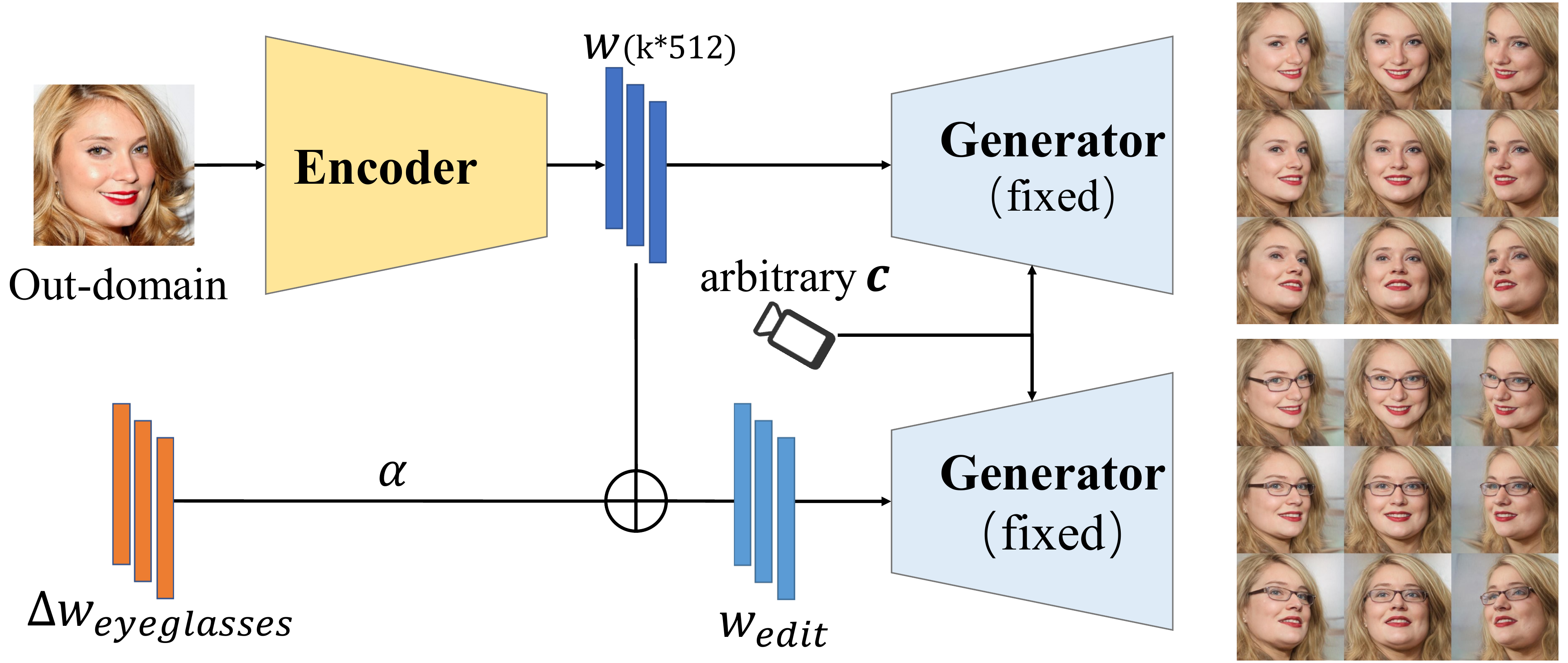}
    \caption{}
    \label{fig:pipeline-c}
  \end{subfigure}
  \vspace{-0.4cm}
  \caption{Pipeline of our 3D consistent inversion encoder and editing in the inversion manifold. (a) The training architecture of the encoder, alternating in-domain and out-domain images. $\mathcal{L}_{w}$ is added in the in-domain round. The multi-view identity loss is calculated between the input image and novel views randomly sampled in the surrounding of the input camera pose. (b) We perform inversions with our encoder on a large real image dataset to produce the inversion manifold $\mathcal{W}_{int}$. Then the editing directions can be found through training an SVM. (c) The inference of 3D GAN inversion and editing. We perform the 3D-aware image editing from a single image conditioned on the desired attribute and arbitrary camera pose.}
  \label{fig:architecture}
\end{figure*}
  
\subsection{Overview}
\label{sec:overview}
Current methods of 3D GAN inversion include encoder-based, optimization-based, and hybrid both.
The optimization and hybrid methods are time-consuming, so they are not suitable for interactive applications such as avatar-based communication. 
Existing encoder-based methods often suffer from low resolution and inferior editability.
To overcome these drawbacks, we introduce an encoder mapping a single real image to the latent space $\mathcal{W}^+$ and suggest performing image attribute editing in the inversion manifold.

Our method is based on a pretrained 3D generator such as EG3D, which can synthesize multi-view images conditioned on camera parameters $c$ and noise code $z \in \mathcal{Z}\subseteq\mathbb{R}^{512} $, where $z \sim \mathcal{N}(0,1)$.
The mapping network of the generator transfers $z$ to an intermediate latent code $w \in \mathcal{W}\subseteq\mathbb{R}^{512} $, where $w$ is a 512-dimensional vector from a distribution without explicit formula.
The generator $G$ takes in camera parameters $c$ and the latent code $w$ replicated $k$ times to synthesize images with the desired resolution as described:
\begin{equation}
  X = G(w, c),
  \label{eq:generator}
\end{equation}
where $c$ denotes the camera parameters, including intrinsics and extrinsics.
These replicated latent codes form the space $\mathcal{W}^+\subseteq\mathbb{R}^{k*512}$.
It has been shown that $k$ different style codes, rather than all the same style code, can increase the representation capacity of the generator~\cite{shen2020interfacegan}.
Therefore, in our work, we use the following encoder $E$ to invert the given image to the latent code $w \in\mathcal{W}^+$:
\begin{equation}
\begin{aligned}
  & w = E(X),
  \\&\hat{X} = G(E(X), c).
  \label{eq:encoder}
\end{aligned}
\end{equation}
Then the fixed generator $G$ takes $w$ and a given $c$ to produce the inversion image $\hat{X}$.

Editing images is walking along the editing directions in the latent space, which can be linear or non-linear, here we consider linear editing formally given by
\begin{equation}
  X_{edit} = G(w + \alpha \Delta w, c),
  \label{eq:editing}
\end{equation}
where $\Delta w$ is the editing direction, $\alpha$ is the editing degree.

In the following, we will present how to improve the 3D consistency of the above encoder $E$ and how to find a more precise editing direction. 
The whole architecture is illustrated in Figure \ref{fig:architecture}.

\subsection{3D Consistent Encoder for Inversion}
\label{sec:encoder}
Despite more fidelity near the input camera pose, optimization and hybrid methods will lead to 3D inconsistency of views at large camera poses due to optimization on a single image.
The encoder methods alleviate this problem by learning the features of a large number of images with different views.
To further improve 3D consistency, we explicitly encourage 3D consistency during training the encoder in two ways: \textbf{alternating training scheme} and \textbf{multi-view identity loss}, as detailed below.

\vspace{2pt}
\noindent
\textbf{Alternating training scheme.}\quad Unlike previous approaches that train the encoder with only real images, we propose an alternating training scheme, which includes in-domain iteration and out-domain iteration.
The model takes in a batch of out-domain images, followed by a batch of in-domain images.
The encoder is optimized to reconstruct the input images when inverting the out-domain images.
When inverting the in-domain images, we additionally optimize the encoder to reconstruct the latent because we have the ground truth. The additional latent code regularization term is
\begin{equation}
  \mathcal{L}_{w}=\mathbb{E}_w\left[\Vert w - E(G(w, c))\Vert_2^2\right],
  \label{eq:loss_reg}
\end{equation}
where $G$ is the pretrained generator and $E$ is the encoder.

Alternating training scheme brings two benefits: (i) The training dataset is augmented, increasing the diversity of contents and poses seen by the model; (ii) The regression of the ground latent code encourages the distribution of the inverted latent code closer to the distribution of the original latent code of the generator.

\vspace{2pt}
\noindent
\textbf{Multi-view identity loss.}\quad
For precise face image inversion and editing, it is challenging to preserve the identity of the input subject.
To tackle this, we impose a specific identity loss~\cite{richardson2021encoding}, defined by 
\begin{equation}
  \mathcal{L}_{ID}(X,c)=1 - \langle F(X), F(G(E(X), c))  \rangle ,
  \label{eq:loss_id}
\end{equation}
where $F$ is the pretrained ArcFace~\cite{JiankangDeng2021ArcFaceAA} network which extracts the feature of face, $\langle\cdot\rangle$ is the cosine similarity.
$\mathcal{L}_{ID}(X,c)$ denotes the identity loss between the input image $X$ and the image generated by the inversion latent code of the image and the given camera pose $c$.

To improve the identity similarity between images with different poses, we propose a novel loss $\mathcal{L}_{multiID}$ including two terms: the identity loss at the original camera pose and the average identity loss at the surrounding camera pose. It is defined by
\begin{equation}
\begin{aligned}
  \mathcal{L}_{multiID}=&\lambda_{ori} \mathcal{L}_{ID}(X,c_{ori}) + 
  \\& \lambda_{sur} \frac{1}{N}\sum_{i=1}^{N} \mathcal{L}_{ID}(X,c_{sur}^i),
  \label{eq:loss_mulID}
\end{aligned}
\end{equation}
where $c_{ori}$ are the camera parameters of the input view, $c_{sur}^i$ are the camera parameters of the views surrounding the input view, $N$ is the number of surrounding views sampled, $\lambda_{ori}$ and $\lambda_{sur}$ are the weights of each loss term, respectively.
Different sampling strategies can be used to sample the surrounding views centered on the input view.
In this paper,  we uniformly sample $N=4$ views from yaw angles between $[-20^{\circ},20^{\circ}]$ and pitch angles between $[-5^{\circ},5^{\circ}]$ for an input image.

\subsection{Total Losses}
\label{sec:loss}
Our encoder can be trained with natural images in an end-to-end manner.
We calculate the commonly used $\mathcal{L}_2$ and $\mathcal{L}_{LPIPS}$~\cite{zhang2018unreasonable} losses between the input image $X$ and the inversion image $\hat{X}$ with the input camera pose to improve the pixel-wise and perceptual similarities.

The total loss is defined as a weight aggregation of all the losses above:
\begin{equation}
\begin{aligned}
  \mathcal{L}=&\lambda_{l2} \mathcal{L}_{2} + \lambda_{lpips} \mathcal{L}_{LPIPS} + \lambda_{w}\mathcal{L}_{w} + \mathcal{L}_{multiID},
  \label{eq:total_loss}
\end{aligned}
\end{equation}
where $\lambda_{l2}$, $\lambda_{lpips}$, and $\lambda_{w}$ are the weights of each loss term, respectively.
Note that $\mathcal{L}_{multiID}$ is only used in the human face domain. 
We simply perform a grid search on these weights to guide the model to produce high-fidelity reconstruction results.

\subsection{Image Editing in the Inversion Manifold}
\label{sec:editing}
\begin{figure}[t]
  \centering
   \includegraphics[width=1.0\linewidth]{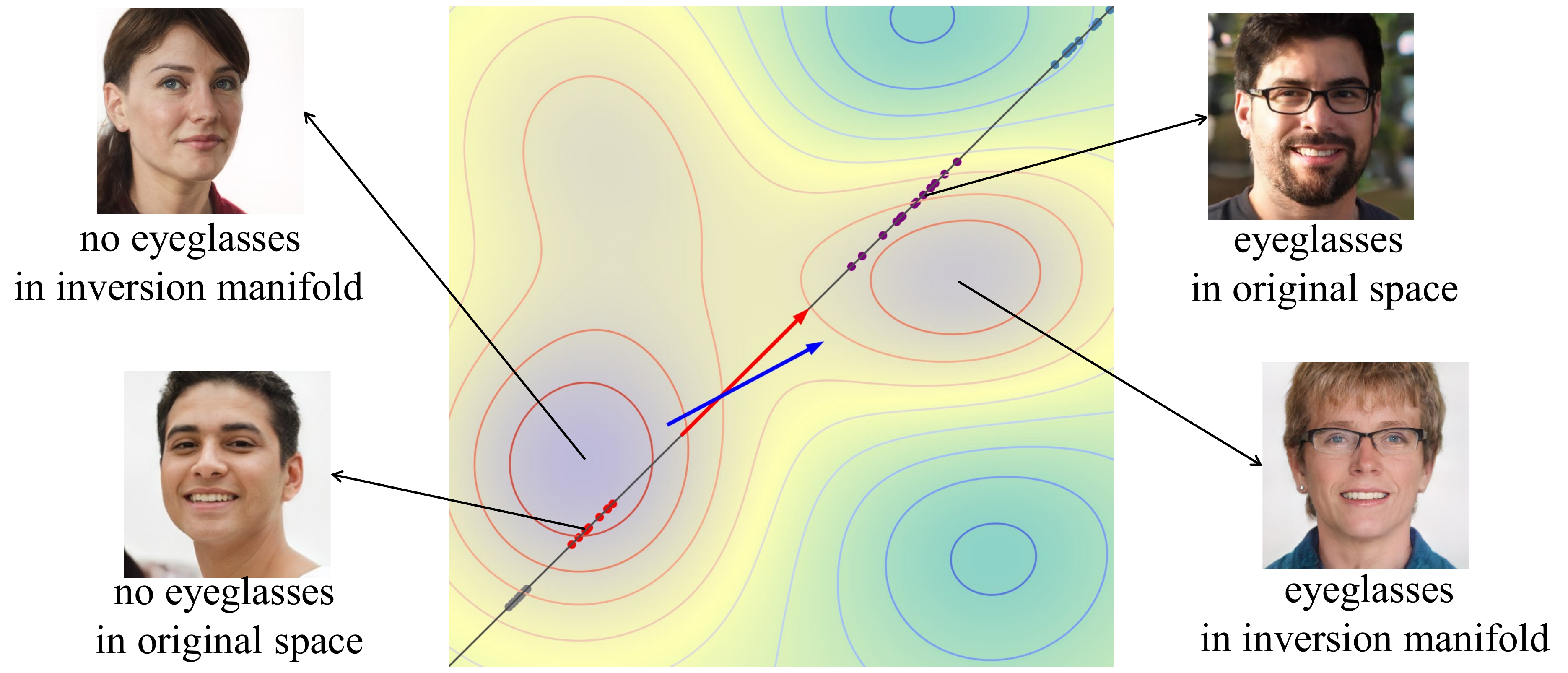}
    \vspace{-0.5cm}
   \caption{The distortion between the original latent space and the inversion manifold for 1-dimensional $\mathcal{W}$ and $k=2$. The points in the diagonal line are sampled from the original latent space $\mathcal{W}_{ori}$, which here is represented as a 1-dimensional Gaussian mixture distribution. The entire region represents $\mathcal{W}^2$ space. The warmer cluster corresponds to a higher density of inversion manifold $\mathcal{W}_{inv}$. The red arrow is the editing direction from no eyeglasses in $\mathcal{W}_{ori}$ to eyeglasses in $\mathcal{W}_{ori}$. The blue arrow is the editing direction from no eyeglasses in $\mathcal{W}_{inv}$ to eyeglasses in $\mathcal{W}_{inv}$.
   }
    \vspace{-0.5cm}
   \label{fig:manifold}
\end{figure}

\begin{figure*}
  \centering
    \includegraphics[width=.98\textwidth]{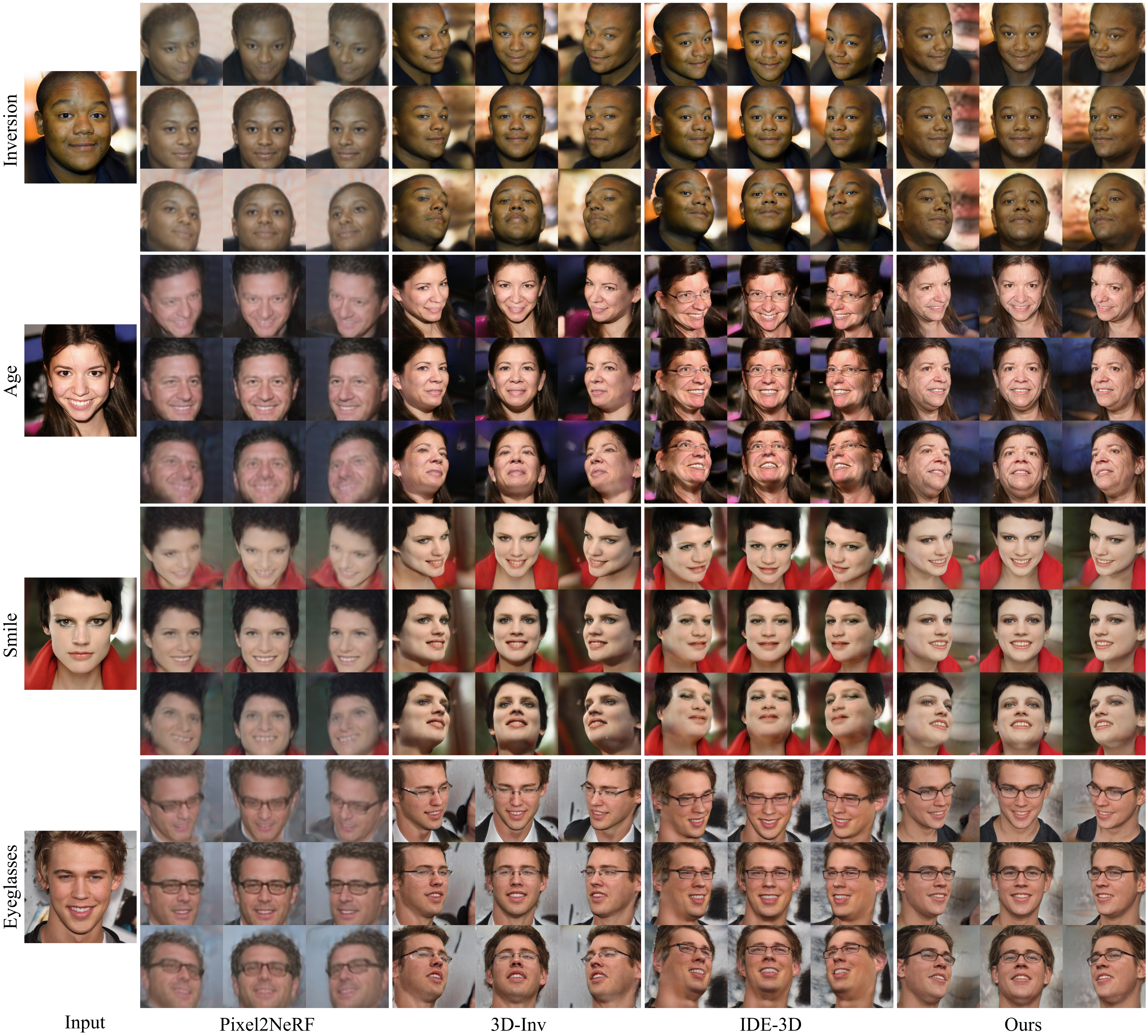}
    \vspace{-0.3cm}
  \caption{Qualitative comparison on face inversion and editing at multiple camera poses.}
  \label{fig:quality_compare}
\end{figure*}

The general approach to attribute-conditional latent code editing is to find a semantic editing direction in the latent space to change the binary labeled attribute (e.g., young $\leftrightarrow$ old, or no smile $\leftrightarrow$ smile)~\cite{shen2020interfacegan,richardson2021encoding,tov2021designing,wang2022high,ling2021editgan,harkonen2020ganspace,shen2021closed}.
Take space $\mathcal{W}^+$ as an example, we sample a latent code $w\in\mathcal{W}^+$, where $\mathcal{W}^+\subseteq\mathbb{R}^{k*512}$.
Formally, we are seeking an editing direction $\Delta w\in\mathcal{W}^+$ such that $w_{edit} = w + \alpha \Delta w$.
$\alpha>0$ will make the edited image look more positive on the attribute, and $\alpha<0$ represents more negative.

To edit the real image, we need to perform inversion as described in Section \ref{sec:overview} to obtain the latent code of the image.
Here, we have multiple candidates for latent space such as $\mathcal{Z}$, $\mathcal{W}$, $\mathcal{W^+}$, and $\mathcal{S}$~\cite{wu2021stylespace}.
Strictly speaking, these spaces are manifolds on the corresponding linear space in mathematical terms.
However, note that for consistency and simplicity, we still refer to them as space.
As described in Section \ref{sec:overview}, we can sample latent codes in these spaces and use them as input to the generator to synthesize the in-domain images.
We refer to these spaces used for in-domain images  as the original latent spaces of the generator, e.g. the $\mathcal{W^+}$ space is denoted as $\mathcal{W}_{origin}^+$.
Instead, we propose that the space consisting of latent codes obtained by inverting a large number of real images via an inversion encoder is called \textit{inversion manifold}.
Each original latent space has a corresponding inversion manifold, e.g. the $\mathcal{W}_{origin}^+$ corresponds to $\mathcal{W}_{inv}^+$.
To avoid confusion, we call it \textit{inversion manifold} instead of \textit{inversion space}.

It's a non-trivial task to invert a real image to latent code because the generator can not fully model the true distribution~\cite{shen2020interfacegan}.
Furthermore, it turns out that there is a distortion between $\mathcal{W}_{inv}^+$ and $\mathcal{W}_{origin}^+$ because the $w'$ obtained by the inversion of the in-domain image generated from $w$ is not equal to $w$.
Current popular editing techniques such as InterfaceGAN~\cite{shen2020interfacegan}, GANSPace~\cite{harkonen2020ganspace},  StyleFlow~\cite{wu2021stylespace}, etc., all learn the editing directions in the original latent space.
Therefore, editing the latent code in $\mathcal{W}_{inv}^+$ with the editing direction found in $\mathcal{W}_{origin}^+$ leads to distortion.
The distortion can be described as:
\begin{equation}
  d(\Delta w) = \Delta w_{inv} - \Delta w_{origin} ,
  \label{eq:n_distortion}
\end{equation}
where $\Delta w_{origin}$ is the editing direction found in $\mathcal{W}_{origin}^+$, $\Delta w_{inv}$ is the editing direction found in $\mathcal{W}_{inv}^+$. 
While $\Delta w_{origin}$ can be used to edit the in-domain image very well, it will lead to imprecision when editing the real image on the inversion manifold by using $\Delta w_{origin}$. 
We present an illustration of the distortion between $\Delta w_{origin}$ and $\Delta w_{inv}$ in Figure \ref{fig:manifold}.
Our method is orthogonal to existing editing techniques, employing them to find editing directions on the inversion manifold.
We show the differences between the results between editing in the original space and inversion manifold in Table \ref{tab:inversion_editing} and Figure \ref{fig:inversion_editing}.

\section{Experiments}
\label{sec:experiments}
\subsection{Experimental Settings}
\noindent
\textbf{Datasets.}\quad
For the human face domain, we train the inversion encoder with FFHQ~\cite{karras2019style} dataset cropped as ~\cite{chan2022efficient} and use CelebA-HQ~\cite{TeroKarras2017ProgressiveGO} for evaluation.
We augment the datasets with horizontal flips and estimate the camera parameters of the images following ~\cite{chan2022efficient}.
We use InterfaceGAN~\cite{shen2020interfacegan} for finding the attribute editing directions.
The implementation details are provided in Appendix A.2.

\subsection{Evaluation}

We compare our method with three state-of-the-art methods for 3D GAN inversion: Pixel2NeRF~\cite{cai2022pix2nerf}, IDE-3D~\cite{sun2022ide}, and 3D-Inv~\cite{lin20223d}.
Note, Pixel2NeRF and our method are both encoder-based methods. IDE-3D is a hybrid method.
3D-Inv is an optimization-based method.
In the comparison experiments, we use the official pretrained models and code for both Pixel2NeRF and IDE-3D, and we implement 3D-Inv according to the paper because they do not release the code.
The metrics are calculated on the first $300$ images from CelebA-HQ. 
Because most of these images are front views,
we uniformly sample $20$ views from yaw angles between $[-30^{\circ},30^{\circ}]$ and pitch angles between $[-20^{\circ},20^{\circ}]$ for a source image.

\begin{table}
  \centering
  \resizebox{0.9\columnwidth}{!}{%
  \begin{tabular}{cccccc}
    \toprule
    Method  & ID  & ID$_{20-30}$ & APD  & Time(s) \\
    \midrule
    IDE-3D~\cite{sun2022ide}         & 0.475 & 0.397 & 0.00139 & 277.5 \\
    3D-Inv~\cite{lin20223d}         & 0.476 & 0.457 & 0.00136 & 238.3  \\
    \midrule
    Pixel2NeRF~\cite{cai2022pix2nerf} & 0.395 & 0.379 & 0.00453 & 0.5  \\
    PREIM3D (Ours)                          & \textbf{0.606} & \textbf{0.576} & \textbf{0.00117} & \textbf{0.05}  \\
    \bottomrule
  \end{tabular}}
    \vspace{-0.3cm}
  \caption{Quantitative evaluation for inversion on faces. ID$_{a-b}$ denotes the mean ArcFace similarity score between the input image and the 20 inverted images uniformly sampled from yaw angles between $[-b^{\circ},a^{\circ}]\cup[a^{\circ},b^{\circ}]$ and pitch angles between $[-20^{\circ},20^{\circ}]$.}
  \label{tab:inversion_compare}
\end{table}

\begin{table}
  \centering
  \resizebox{0.9\columnwidth}{!}{%
  \begin{tabular}{cccccc}
    \toprule
    Metric & Method & Age & Smile & Eyeglasses \\
    \midrule
    { \multirow{3}*{ID} } 
    &IDE-3D~\cite{sun2022ide}         & 0.344 & 0.427 & 0.346  \\
    &3D-Inv~\cite{lin20223d}         & 0.425 & 0.482 & 0.420  \\
    &Pixel2NeRF~\cite{cai2022pix2nerf} & 0.219 & 0.324 & 0.262  \\
    &PREIM3D (Ours)                          & \textbf{0.557} & \textbf{0.614} & \textbf{0.531}  \\
    \midrule
    { \multirow{3}*{AA} } 
    &IDE-3D~\cite{sun2022ide}         & 1.35 & 1.41 & 1.52  \\
    &3D-Inv~\cite{lin20223d}         & 1.41 & 1.49 & 1.61  \\
    &Pixel2NeRF~\cite{cai2022pix2nerf} & 1.42 & 1.47 & 1.44  \\
    &PREIM3D (Ours)                          & \textbf{1.51} & \textbf{1.54} & \textbf{1.62}  \\
    \midrule
    { \multirow{3}*{AD} } 
    &IDE-3D~\cite{sun2022ide}         & 1.04 & 0.56 & 0.74  \\
    &3D-Inv~\cite{lin20223d}         & 0.94 & 0.56 & 0.62   \\
    &Pixel2NeRF~\cite{cai2022pix2nerf} & 1.23 &0.57 & 0.79  \\
    &PREIM3D (Ours)                         & \textbf{0.82} & \textbf{0.49} & \textbf{0.50}   \\
    \bottomrule
  \end{tabular}}
    \vspace{-0.3cm}
  \caption{Quantitative evaluation for attribute editing on faces. Attribute altering (AA) measures the change of the desired attribut. Attribute dependency (AD) measures the degree of change on other attributes when edit a certain attribute.}
  \label{tab:edit_compare}
\end{table}

\noindent
\textbf{Quantitative Evaluation.}\quad 
Table \ref{tab:inversion_compare} provides quantitative comparisons of the 3D GAN inversion performance.
We measure multi-view facial identity consistency (ID) with the average ArcFace similarity score~\cite{JiankangDeng2021ArcFaceAA} between the sampled images and the source image.
Pose accuracy is evaluated by the average pose distance(APD), which is root mean squared error between the pose encodings estimated by the pretrained 3D face detector~\cite{JiankangDeng2021ArcFaceAA}.
Time metric indicates the average inference time (encoding time and generation time) for one image computed on one Tesla V100 GPU.
Our method outperforms baselines on ID and APD metrics and is significantly faster than IDE-3D and 3D-Inv when inference.

We show the comparison of face attribute editing against the baselines in Table \ref{tab:edit_compare}.
We use an off-the-shelf multi-label classifier based on ResNet50~\cite{DBLP:conf/iccv/HuangB17} to obtain predicted logits.
Attribute altering (AA) measures the change of the desired attribute, which is the attribute logit change $\Delta l_t$~\cite{wu2021stylespace} when detecting attribute $t$ by the classifier (pretrained on CelebA~\cite{liu2015faceattributes}).
$\Delta l_t$ is normalized by $\sigma (l)$~\cite{wu2021stylespace}, which is the standard deviation calculated from the logits of CelebA-HQ dataset.
We evaluate the precision of attribute editing with attribute dependency (AD)~\cite{wu2021stylespace}, which measures the degree of change on other attributes when modifying along a certain attribute editing direction, as measured by classifiers.
Our method performs better than the previous method. 
More attribute editing results are provided in Appendix.

\vspace{2pt}
\noindent
\textbf{Qualitative Evaluation.}\quad
We present examples of the inversion and editing results in Figure \ref{fig:quality_compare}. 
We sample $9$ images for each source image with $yaw=[-30^{\circ}, 0, 30^{\circ}]$ and $pitch=[-20^{\circ}, 0, 20^{\circ}]$.
While optimization-based methods perform better near the input camera pose, optimization on a single image produces artifacts at large camera pose changes, such as head deformation. 
Compared with previous methods, our method achieves the best 3D consistency, especially at large camera pose. 
We provide more examples in the Appendix, not only for human faces but also for cats.

\subsection{Ablation Study}
We conduct an ablation study to further validate the benefits of our proposed components and strategies.

\begin{figure}
  \centering
  \includegraphics[width=0.98\linewidth]{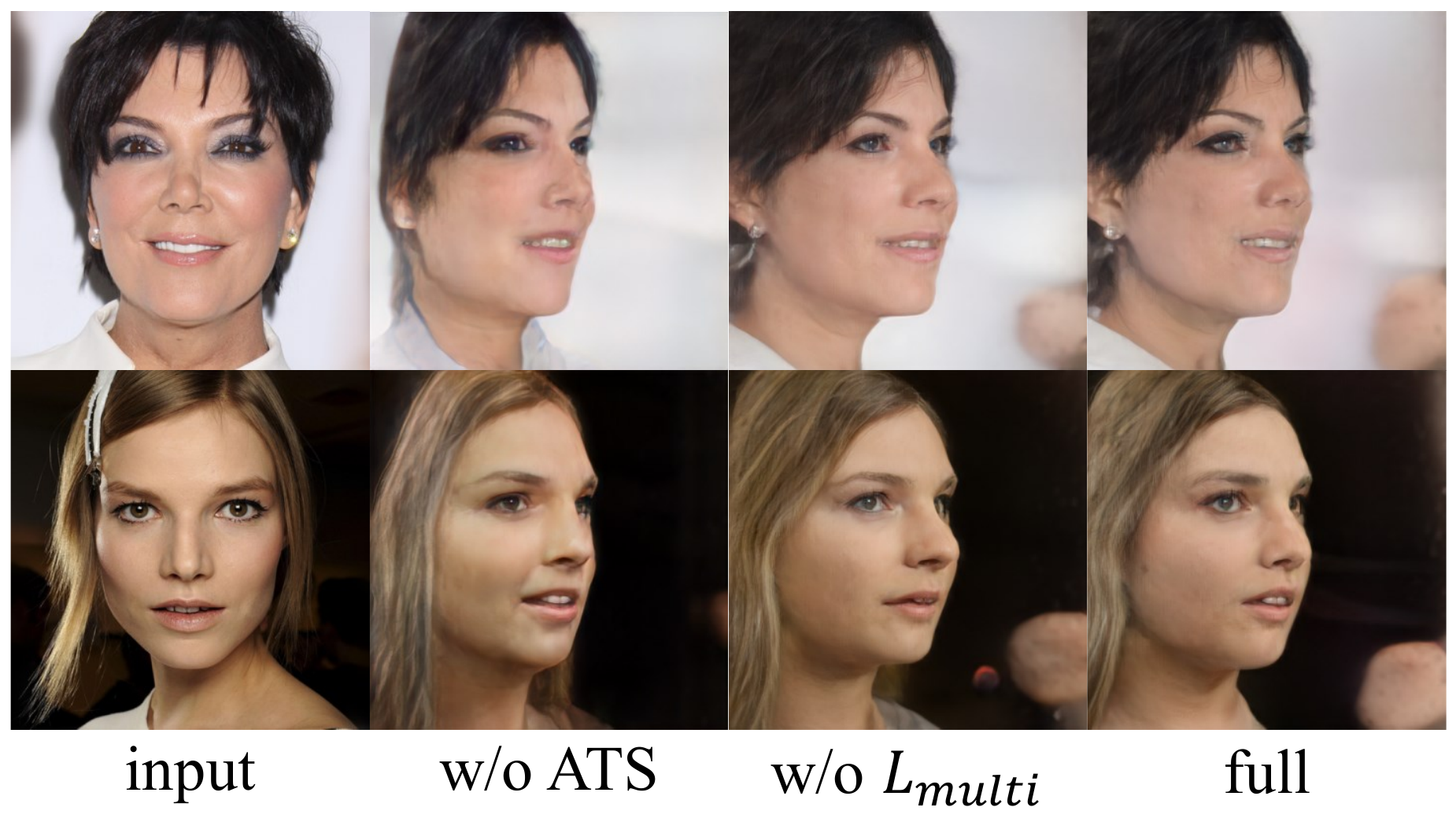}
    \vspace{-0.3cm}
   \caption{Inversion results at large camera pose. w/o ATS denotes the model without an alternating training scheme. w/o $\mathcal{L}_{multiID}$ denotes the model without multi-view ID loss.
   }
   \label{fig:ablation_reg}
\end{figure}
\noindent
\textbf{Alternating training scheme.}\quad
As analyzed before, the alternating training scheme and the ground latent code regression encourage the inverted latent space to match the original latent space of the generator.
We believe that this will better maintain the 3D consistency of the 3D generator.
To validate the effectiveness of this strategy, we show the inversion results in Figure \ref{fig:ablation_reg}.
Without the alternating training scheme and the ground latent code regression, the model will lead to significant 3D inconsistency.

\begin{table}
  \centering
  \resizebox{0.9\columnwidth}{!}{%
  \begin{tabular}{ccccc}
    \toprule
     & ID$_{0-10}$ & ID$_{10-20}$ &  ID$_{20-30}$ & ID$_{r}$ \\ 
    \midrule
    w/o ATS         &0.517  & 0.507  &0.482  & 0.503 \\
    w/o $\mathcal{L}_{multiID}$         &0.572  & 0.563  &0.541  & 0.557 \\
    full          & 0.629  &0.611  & 0.576  & 0.606\\
    \bottomrule
  \end{tabular}}
    \vspace{-0.2cm}
  \caption{Effects of alternating training scheme and multi-view ID loss.}
  \label{tab:ablation_multiview}
\end{table}
\noindent
\textbf{Multi-view ID Loss.}\quad
For the human face domain, the multi-view ID loss explicitly guides the model to preserve the identity of the input subject.
Following EG3D~\cite{chan2022efficient}, we calculate the mean ArcFace similarity score between images of the same inverted face at two random camera pose.
Our inversion model scored 0.82, while the pretrained EG3D scored 0.77 as reported in their paper.
As shown in Table \ref{tab:ablation_multiview}, the model with multi-view ID loss improves the identity consistency score.
As shown in Figure \ref{fig:ablation_reg}, the multi-view ID loss will encourage subtle face shape adjustments

\begin{table}
  \centering
  \resizebox{0.9\columnwidth}{!}{%
  \begin{tabular}{ccc|cc}
    \specialrule{0.75pt}{0pt}{0pt}
     { \multirow{2}*{Method} }& \multicolumn{2}{c}{AA} &\multicolumn{2}{c}{AD} \\
    \cmidrule(ll){2-5} 
    & origin & inversion & origin & inversion \\
    \specialrule{0.5pt}{0pt}{0pt}
    e4e (2D)~\cite{tov2021designing}         & 1.53 & \textbf{1.59}  &0.45 & \textbf{0.38}  \\
    PREIM3D (Ours)                         & 1.49 & \textbf{1.56} & 0.81 & \textbf{0.60}  \\
    \specialrule{0.75pt}{0pt}{0pt}
    \end{tabular}}
    \vspace{-0.3cm}
    \caption{Performance comparisons in terms of average AA and AD metrics (age, smile, and eyeglasses) on editing in the original space and inversion manifold.}
    \vspace{-0.3cm}
  \label{tab:inversion_editing}
\end{table}

\begin{figure}
  \centering
  \includegraphics[width=0.98\linewidth]{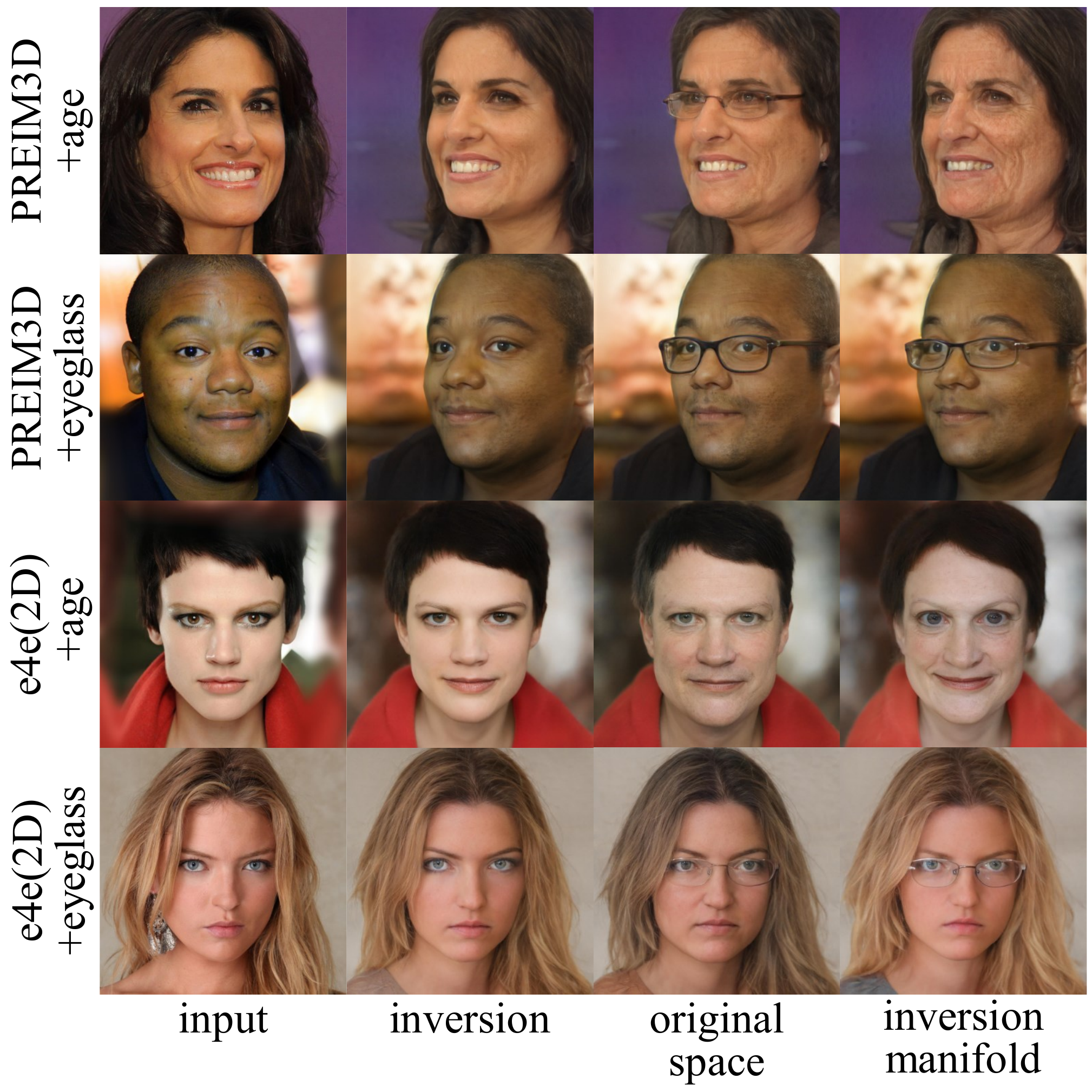}
    \vspace{-0.3cm}
   \caption{Visual comparison between editing in the original space and editing in the inversion manifold.
   }
   \label{fig:inversion_editing}
\end{figure}
\noindent
\textbf{Editing in the Inversion Manifold.}\quad
Similar to 3D GAN inversion, we can also perform editing in the inversion manifold in 2D space.
To validate the generalization of the inversion manifold, we apply it to the state-of-the-art 2D GAN inversion encoder, e4e~\cite{tov2021designing}. 
Table \ref{tab:inversion_editing} shows the improvement of the quantitative metrics.
Figure \ref{fig:inversion_editing} demonstrate editing in the inversion manifold will produce more precise results.
For example, the third column of the second row in Figure~\ref{fig:inversion_editing} shows that editing eyeglasses in the original space will increase beard and gray hair.

\subsection{User Study}
\begin{table}
  \centering
  \resizebox{0.9\columnwidth}{!}{%
  \begin{tabular}{ccccc}
    \toprule
     & IDE-3D & 3D-Inv &  Pixel2NeRF & Origin \\ 
    \midrule
    Inversion         &0.55  & 0.51  &1.0  & - \\
    Editing         & 0.65  &0.89  & 0.98  & 0.70\\
    \bottomrule
  \end{tabular}}
    \vspace{-0.3cm}
  \caption{The result of our user study. The value represents the rate of Ours \textgreater others. Origin indicates editing in the original space using our inversion encoder.}
  \label{tab:userstudy}
\end{table}
Considering the human evaluation, we conduct a user study.
We collect 1,500 votes from 25 volunteers, who evaluate the 3D consistency and realism of the inversion and editing results.
Each volunteer is given a source image, 9 images of our method, and 9 images of baseline (as in Figure ~\ref{fig:quality_compare}).
According to Table \ref{tab:userstudy}, the user study shows our method outperforms the baselines.

\section{Conclusions}
In this paper, we propose a fidelity 3D consistent pipeline that enables 3D reconstruction and 3D-aware editing from a single real image efficiently.
With the alternating training scheme, we perform latent code regression to close the gap between the inversion latent code distribution and the original latent code distribution of the generator.
This scheme leverages the 3D prior information of the generator and helps to maintain 3D consistency.
Benefiting from the multi-view ID loss, our method achieves better identity consistency in the human face domain.
We show that editing in the inversion manifold produces more precise results than in the original latent space.
Our method can be used for many interactive 3D applications such as virtual reality, metaverse, and avatar-based communication.

\noindent
\textbf{Limitations.}\quad
One limitation of our work is the difficulty in dealing with uncommon cases such as delicate earrings and special hairstyles.
As our inversion encoder relies on the capacity of the generator to capture real-world scenes, some details were reconstructed imperfectly.

\clearpage
\appendix

\section*{Appendix}

In the supplement, we first provide implementation details, including encoder training process and edit directions seeking. We follow with additional experiments and visual results. We highly recommend watching the supplemental video, which contains a live demonstration of the real-time inversion and attribute editing and a demonstration of sequential editing synthesis.

\section{Implementation Details}

\subsection{Encoder Training}

We implemented our encoder training on top of the official pSp~\cite{richardson2021encoding} encoder training framework implementation.
We set the $\lambda_{l2}=1.0$, $\lambda_{lpips}=0.8$, and $\lambda_{ori}=0.4$ in the first $20,000$ training steps. After the $20,000$ steps, we gradually add a delta for   the $\lambda_{w}=1e^{-4}$ every $5,000$ steps. After the $100,000$ steps, we gradually add a delta for the $\lambda_{sur}=1e^{-4}$ every $5,000$ steps.
The in-domain images are sampled from yaw angles between $[-30^{\circ},30^{\circ}]$ and pitch angles between $[-20^{\circ},20^{\circ}]$.
The surrounding images are sampled from yaw angles between $[-20^{\circ},20^{\circ}]$ and pitch angles between $[-5^{\circ},5^{\circ}]$ 

\subsection{Edit Directions Seeking}
We use InterfaceGAN~\cite{shen2020interfacegan} to train a SVM to find out the attribute editing directions.
For the editing directions in the original space, the generator is applied to produce $140,000$ images.
For the editing directions in the inversion manifold, we perform inversion with our encoder on FFHQ~\cite{karras2019style} dataset.
Here, we have obtained the latent code $w$ and image pairs.
An off-the-shelf multi-label classifier based on ResNet50~\cite{DBLP:conf/iccv/HuangB17} is applied to predict the images.
We train the SVM (\url{https://github.com/clementapa/CelebFaces_Attributes_Classification/}) to find the hyperplane that distinguishes binary attributes using the latent code $w$ and the corresponding classification result as input.
The normal vector of the hyperplane is the attribute editing direction.

\section{Comparison on Face Inversion at More Camera Poses }

\begin{figure}[t]
  \centering
   \includegraphics[width=1.0\linewidth]{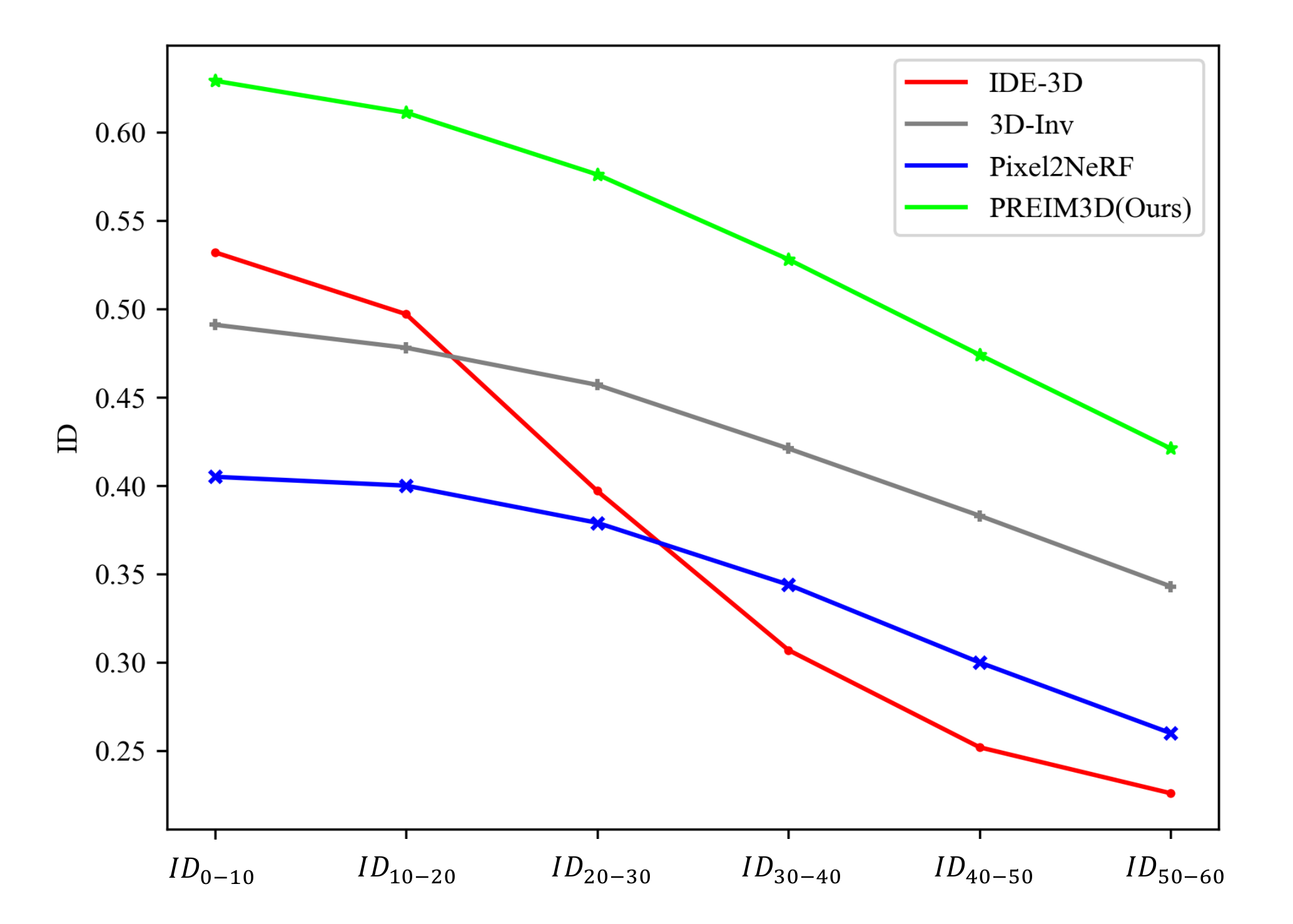}
   \caption{ID$_{a-b}$ denotes the mean ArcFace similarity score between the input image and the 20 inverted images uniformly sampled from yaw angles between $[-b^{\circ},a^{\circ}]\cup[a^{\circ},b^{\circ}]$ and pitch angles between $[-20^{\circ},20^{\circ}]$. Our method has a higher ID score than other methods in different yaw ranges.
   }
   \label{fig:compare_id_plot}
\end{figure}
We uniformly sample 20 inverted images for each image of the first $300$ images from CelebA-HQ in different yaws ranges using IDE-3D, 3D-Inv, Pixel2NeRF, and PREIM3D. 
As with the main text, IDE-3D and 3D-Inv perform image inversion with 500 $w$ optimization steps and 100 generator fine-tuning steps.
We show the identity consistency (ID) in Figure~\ref{fig:compare_id_plot}.

\section{Additional Precise Editing}
\begin{figure*}[t]
  \centering
   \includegraphics[width=1.0\linewidth]{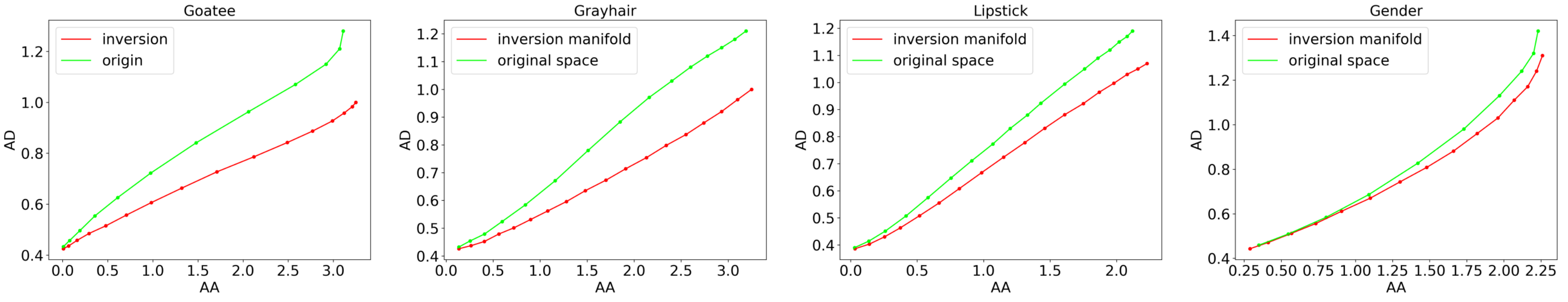}
   \caption{PREIMD(Ours). As the degree of editing $\alpha$ changes, both Attribute Altering (AA) and Attribute Dependency (AD) change. Lower AD indicates more precise.
   }
   \label{fig:precise_editing1}
\end{figure*}

\begin{figure*}[t]
  \centering
   \includegraphics[width=1.0\linewidth]{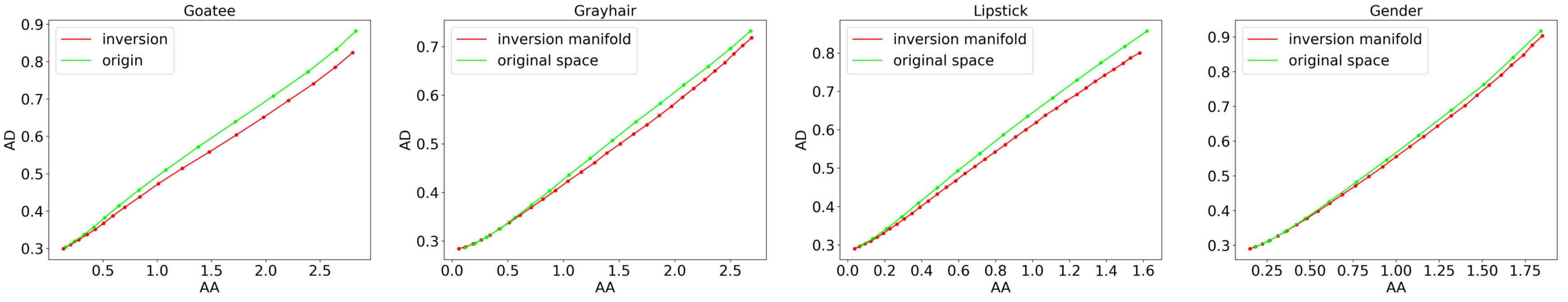}
   \caption{e4e(2D)~\cite{tov2021designing}. As the degree of editing $\alpha$ changes, both Attribute Altering (AA) and Attribute Dependency (AD) change. Lower AD indicates more precise.
   }
   \label{fig:precise_editing2}
\end{figure*}

\noindent
\textbf{AA \& AD.}\quad
Following ~\cite{wu2021stylespace}, we use attribute altering (AA) to evaluate the change of the desired attribute and attribute dependency (AD) to measure the degree of change on other attributes when modifying one attribute.
AA is the change on the logit $\Delta l_t$ of the off-the-shelf multi-label classifier detecting attribute $t$ and is normalized by $\sigma (l_t)$, which is the standard deviation calculated from the logits of CelebA-HQ dataset.
AD measures the change of logit $\Delta l_i$ for other attributes $\forall i \in \mathcal{A}\setminus t$, where $\mathcal{A}$ is the set of all attributes.
Here, we use the mean-AD, defined as $\mathbb{E}(\frac{1}{k}\sum_{i \in \mathcal{A} \setminus t}(\frac{\Delta l_t}{\sigma (l_i)}))$.

To further validate the precision of the editing in the inversion manifold, we perform more attribute editing.
We make different degrees of editing by adjusting $\alpha$, and then observe the changes on the other attributes.
Figure~\ref{fig:precise_editing1},~\ref{fig:precise_editing2} shows the difference between editing in the original space and editing in the inversion manifold, involving goatee, lipstick gray hair, wavy hair, and gender attributes.
Both 2D-space and 3D-space attribute editing show more precise editing in the inversion manifold than in the original space.

\section{Naive Optimization-based Inversion}
\begin{figure}[t]
  \centering
  \includegraphics[width=1.0\linewidth]{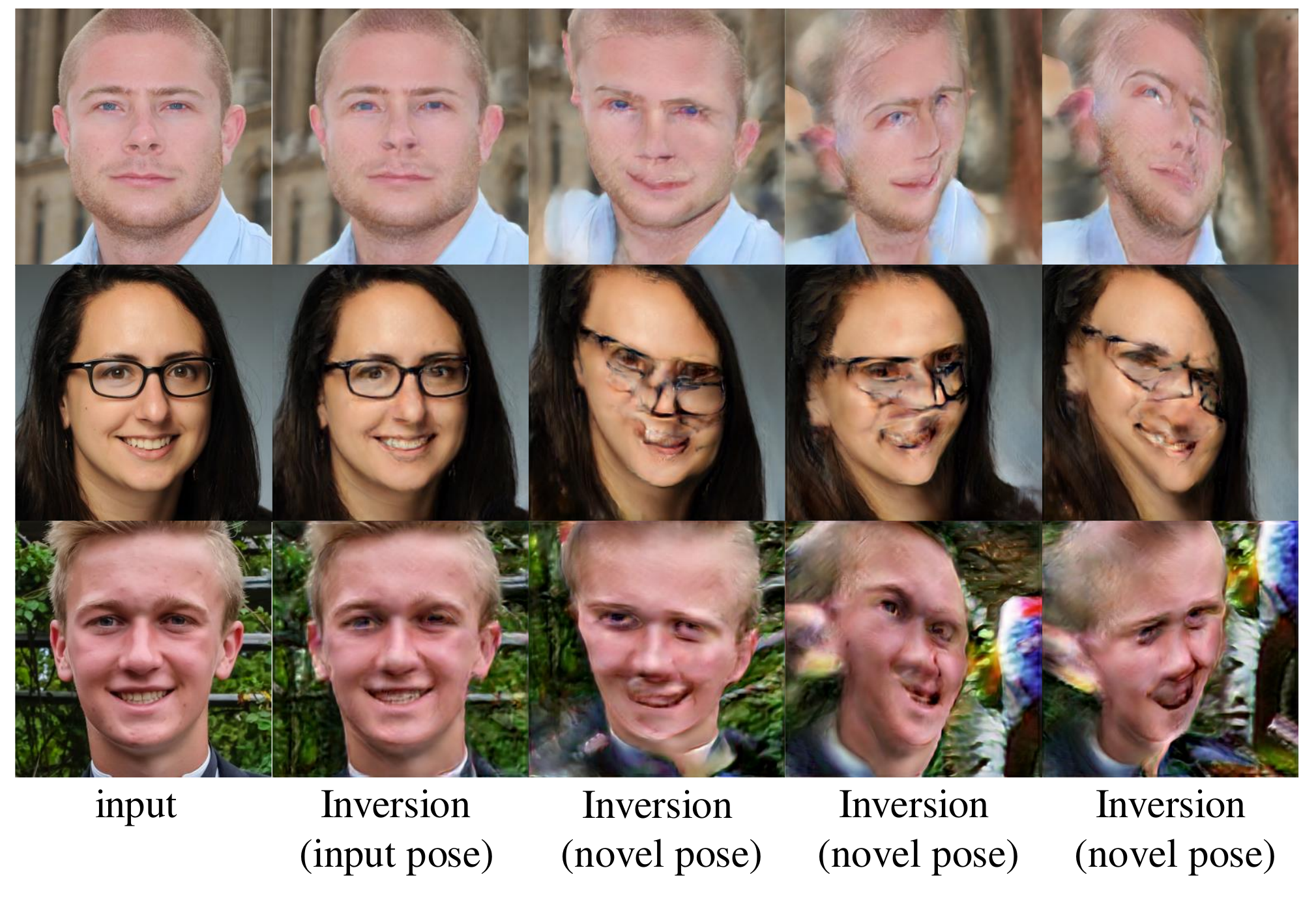}
    \vspace{-0.2cm}
   \caption{The inversion result of $1,0000$ iterations of steps. The naive optimization-based inversion method reconstructs the view of the input camera pose but produces significant artifacts in the views of other camera poses.
   }
    \vspace{-0.2cm}
   \label{fig:optimazation_inversion}
\end{figure}
Different from the PTI technique, the naive optimization-based inversion method only optimizes the latent code $w$, while fixing the generator.
Figure~\ref{fig:optimazation_inversion} shows the inversion results of the naive optimization-based inversion method.

\section{Fine-tuning the Generator}

\begin{figure}[t]
  \centering
  \includegraphics[width=1.0\linewidth]{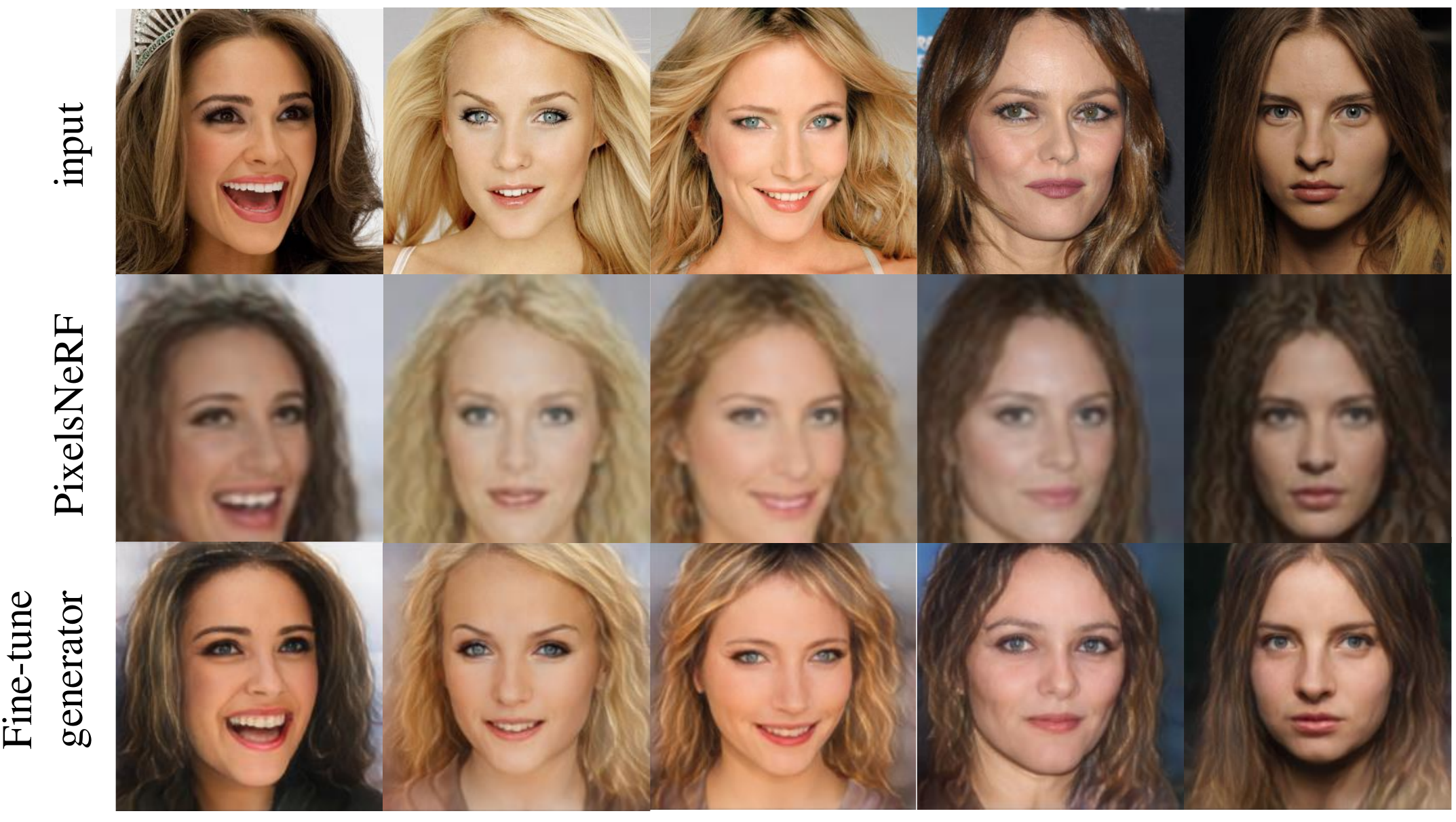}
    \vspace{-0.2cm}
   \caption{It is complex to train the encoder and fine-tune the generator at the same time. We found some tough ripple-like artifacts in the hair.
   }
    \vspace{-0.2cm}
   \label{fig:finetune_generator}
\end{figure}
Inspired by Pixel2NeRF~\cite{cai2022pix2nerf}, we attempted to fine-tune the generator when training the inversion encoder.
Unfortunately, there are always some ripple-like artifacts in the hair, which was also observed for Pixel2NeRF, as shown in the figure~\ref{fig:finetune_generator}.

\section{Beyond Human Face}
\begin{figure}[ht]
  \centering
   \includegraphics[width=1.0\linewidth]{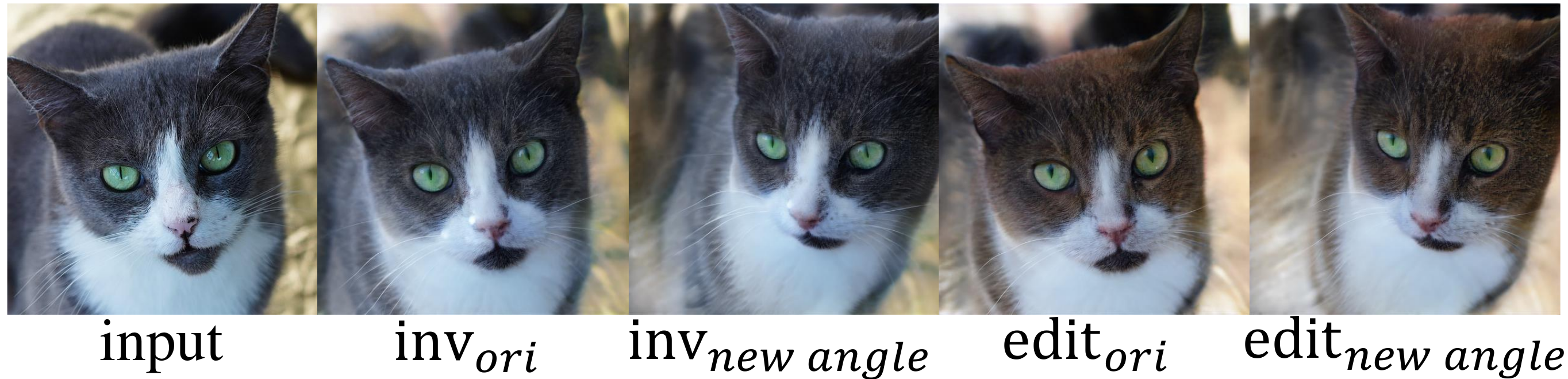}
   \caption{Inversion and (hair color) editing results on cat faces. 
   }
   \label{fig:cat}
\end{figure}
We conducted some experiments with the AFHQ Cat. We invert the dataset to obtain inversion latent samples. 
Following GANSpace, We adopt principal component analysis (PCA) to find the semantic directions. The results in the cat domain are shown in Fig \ref{fig:cat}.

\section{FID and KID}
\begin{table}[h]
  \centering
  \resizebox{0.95\columnwidth}{!}{%
  \begin{tabular}{cccccc}
    \toprule
    Method  & FID$_{ori}$  & FID$_{sm}$ & FID$_{mid}$  & FID$_{la}$ & KID$_{la}$ \\
    \midrule
    IDE-3D        & \textbf{22.7} & \textbf{36.8} & 45.2 & 75.7 & 0.065 \\
    3D-Inv         & 28.1 & 40.6 & \textbf{44.9} & 65.4  & 0.046\\
    Pixel2NeRF & 83.3 & 85.4 & 86.2 & 93.2  & 0.086\\
    PREIM3D (Ours)                          & 43.6 & 48.3 & 50.7 & \textbf{63.3}  & \textbf{0.042}\\
    \bottomrule
  \end{tabular}}
  \caption{FID \& KID comparisons on 1,000 faces from CelebA-HQ. 
  FID$_{ori}$ is measured between the inverted images at the original angle and the input images. 
  We use ${sm}$, ${mid}$, ${la}$ for uniform samples from yaw $[15^{\circ},20^{\circ}]$ and pitch $[10^{\circ},15^{\circ}]$, yaw $[25^{\circ},30^{\circ}]$ and pitch $[15^{\circ},20^{\circ}]$, yaw $[35^{\circ},40^{\circ}]$ and pitch $[20^{\circ},25^{\circ}]$.
  } 
  \label{tab:fid}
\end{table}
We evaluated inversion FID in Table \ref{tab:fid}. 
The inception features used in FID focus on the whole image, while our method introduces regularization of the face regions, which makes our FID scores not as good as IDE-3D and 3D-Inv at small angles. 
However, our model outperforms previous works at large angles. KID shows similar results.

\section{Additional Visual Results}

\begin{figure*}[t]
  \centering
   \includegraphics[width=1.0\linewidth]{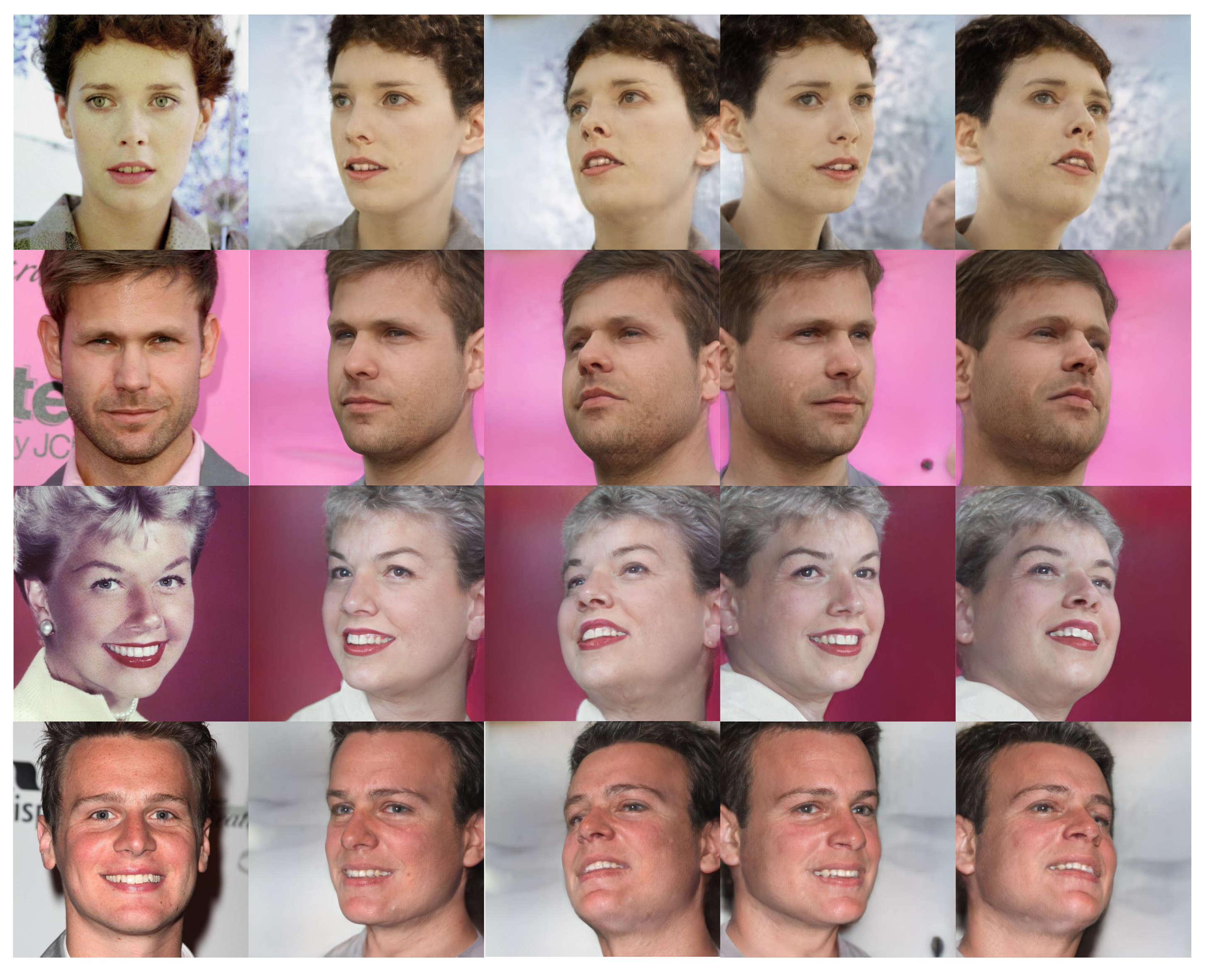}
   \caption{The inversion results obtained by PREIM3D.
   The first column is the input image.
   }
   \label{fig:vr_inversion}
\end{figure*}

\begin{figure*}[t]
  \centering
   \includegraphics[width=1.0\linewidth]{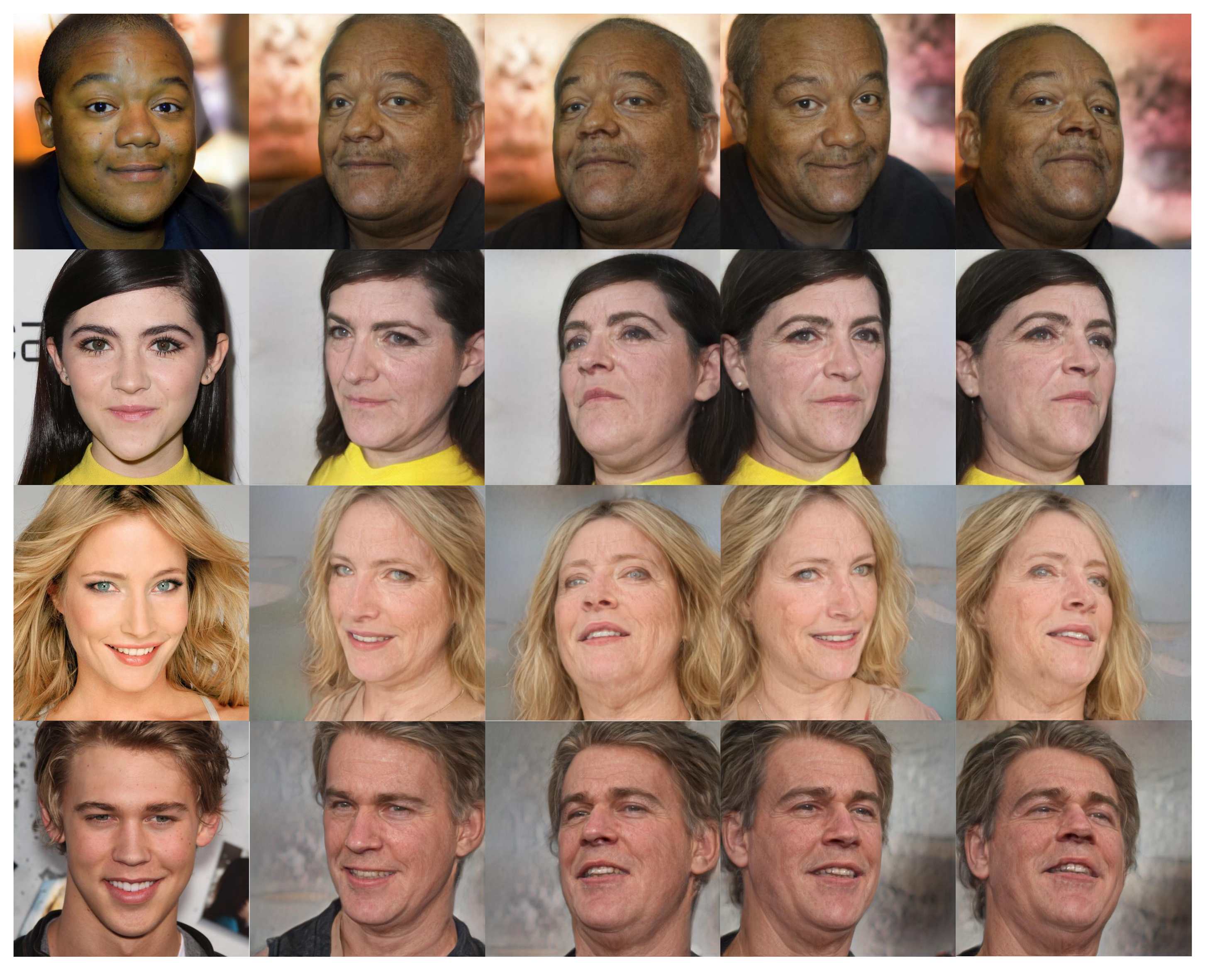}
   \caption{The age editing results obtained by PREIM3D.
   The first column is the input image.
   }
   \label{fig:vr_age}
\end{figure*}

\begin{figure*}[t]
  \centering
   \includegraphics[width=1.0\linewidth]{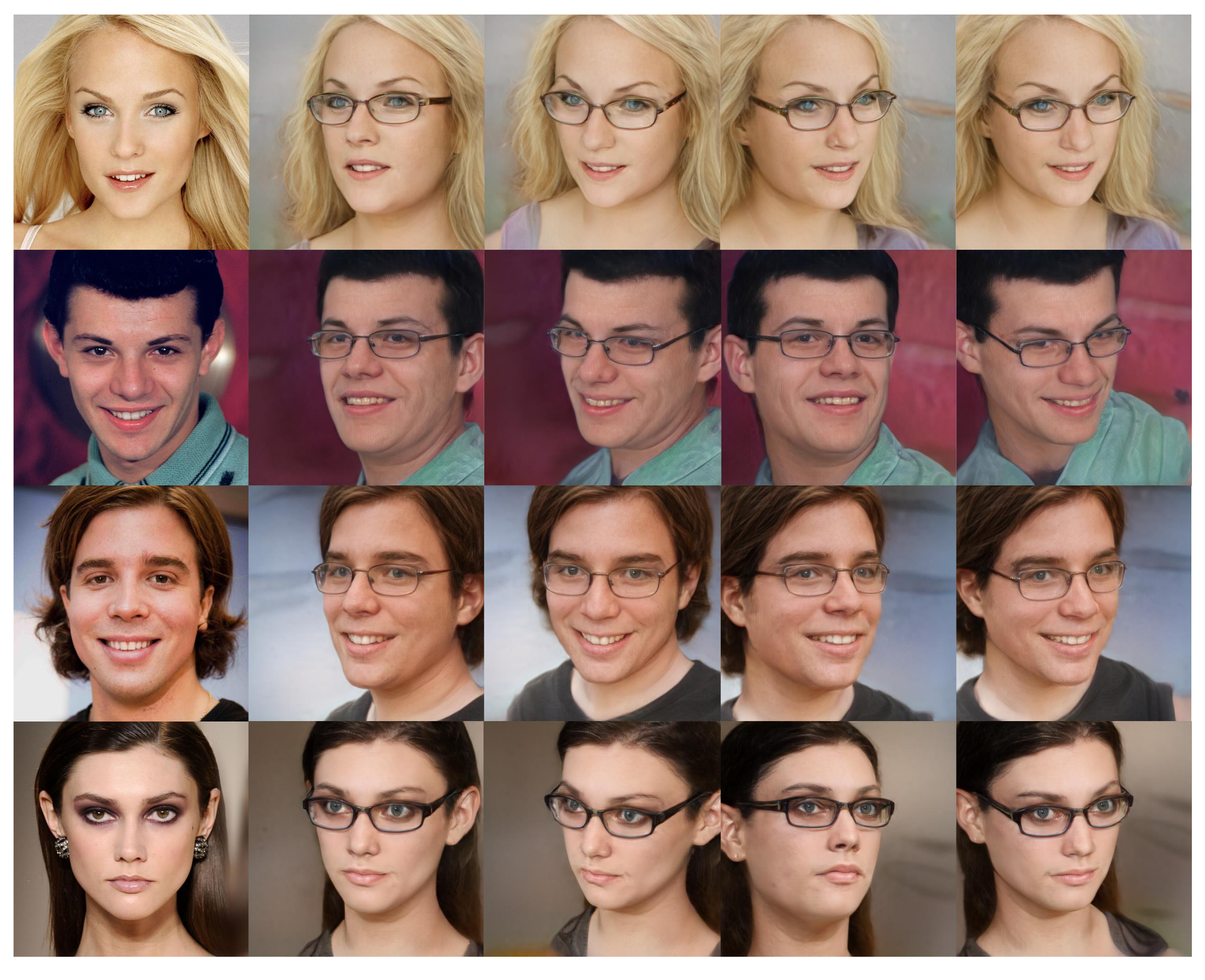}
   \caption{The eyeglasses editing results obtained by PREIM3D.
   The first column is the input image.
   }
   \label{fig:vr_eyeglasses}
\end{figure*}

\begin{figure*}[t]
  \centering
   \includegraphics[width=1.0\linewidth]{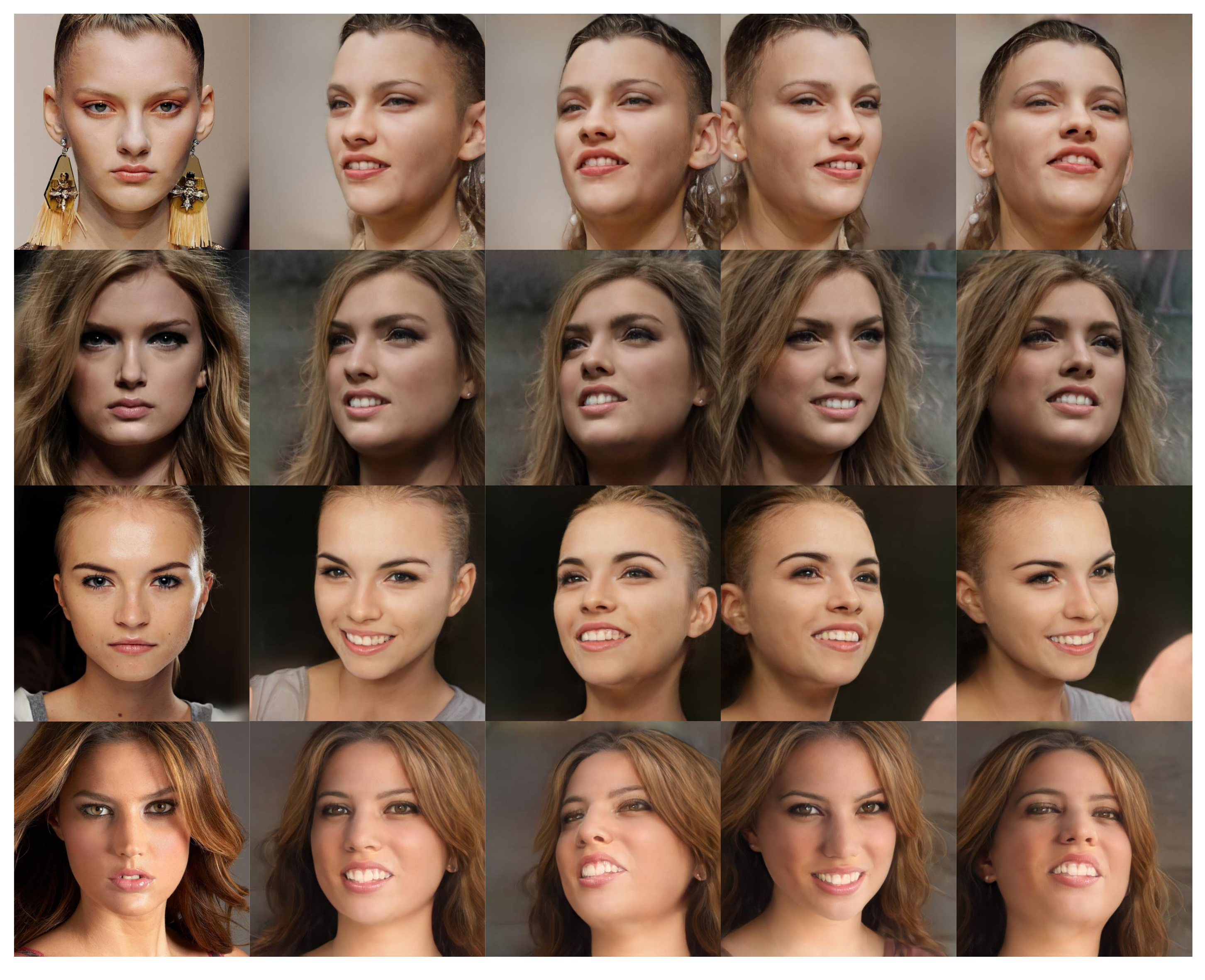}
   \caption{The smile editing results obtained by PREIM3D.
   The first column is the input image.
   }
   \label{fig:vr_smile}
\end{figure*}

\begin{figure*}[t]
  \centering
   \includegraphics[width=1.0\linewidth]{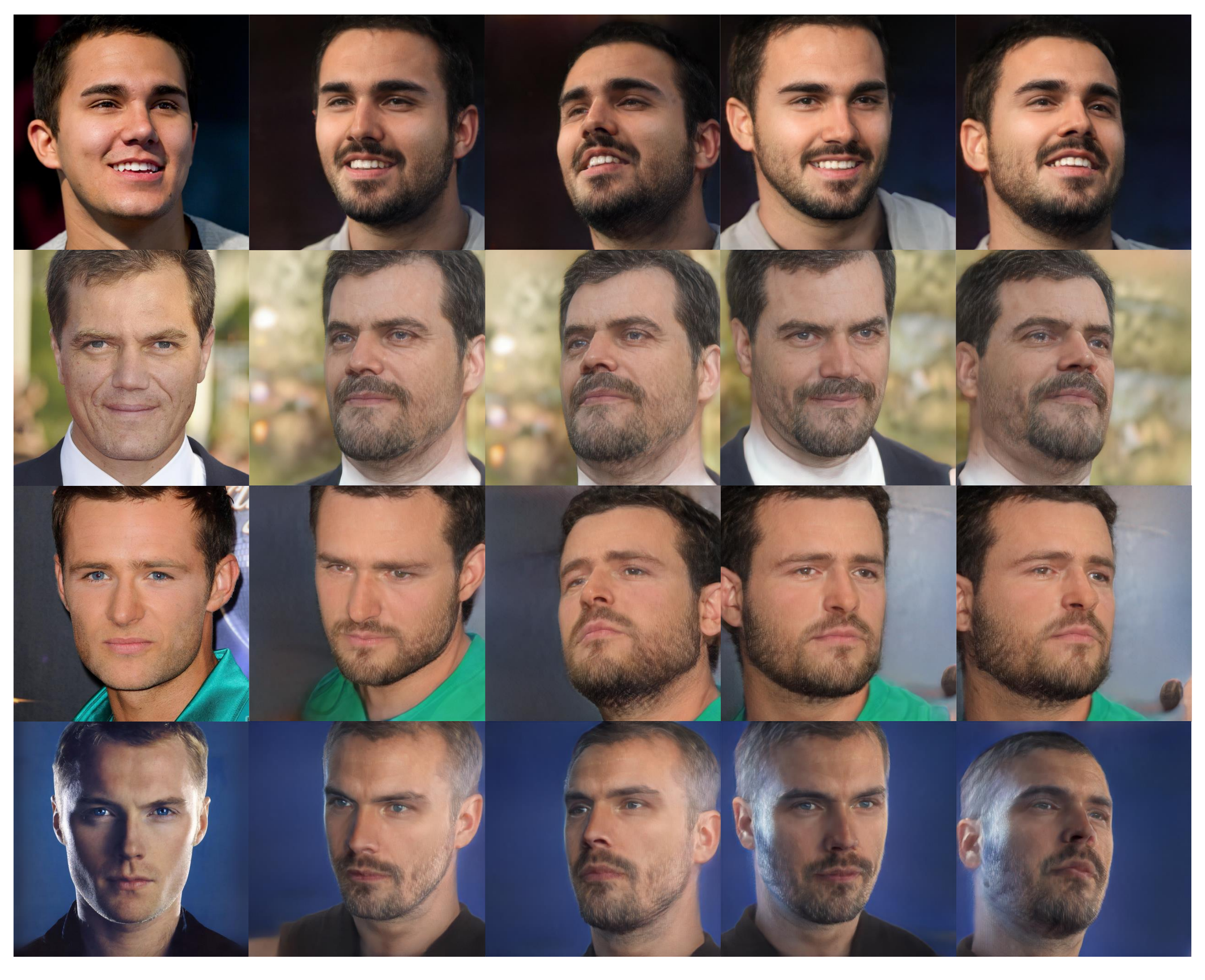}
   \caption{The goatee editing results obtained by PREIM3D.
   The first column is the input image.
   }
   \label{fig:vr_goatee}
\end{figure*}

\begin{figure*}[t]
  \centering
   \includegraphics[width=1.0\linewidth]{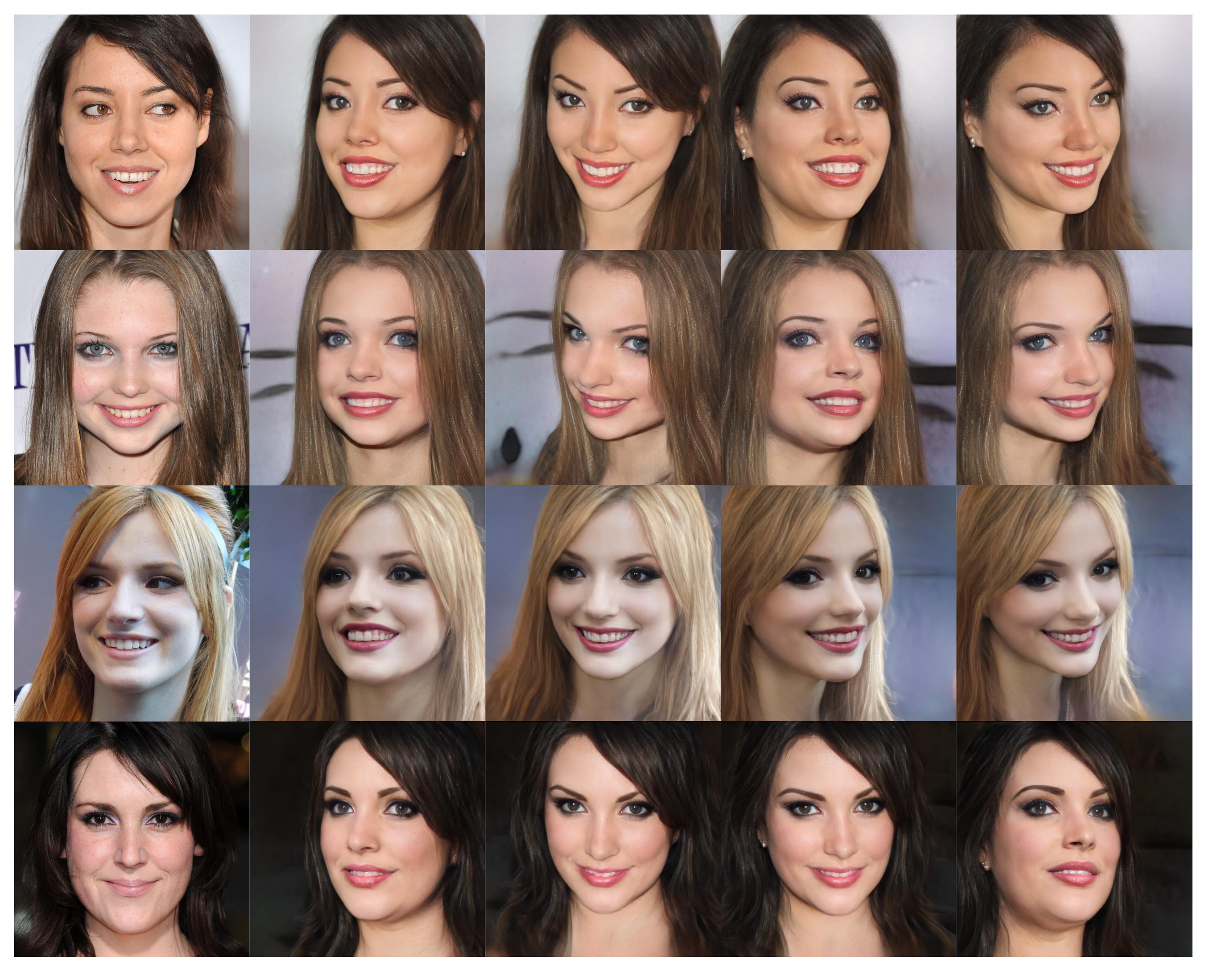}
   \caption{The lipstick editing results obtained by PREIM3D.
   The first column is the input image.
   }
   \label{fig:vr_lipstick}
\end{figure*}

\begin{figure*}[t]
  \centering
   \includegraphics[width=1.0\linewidth]{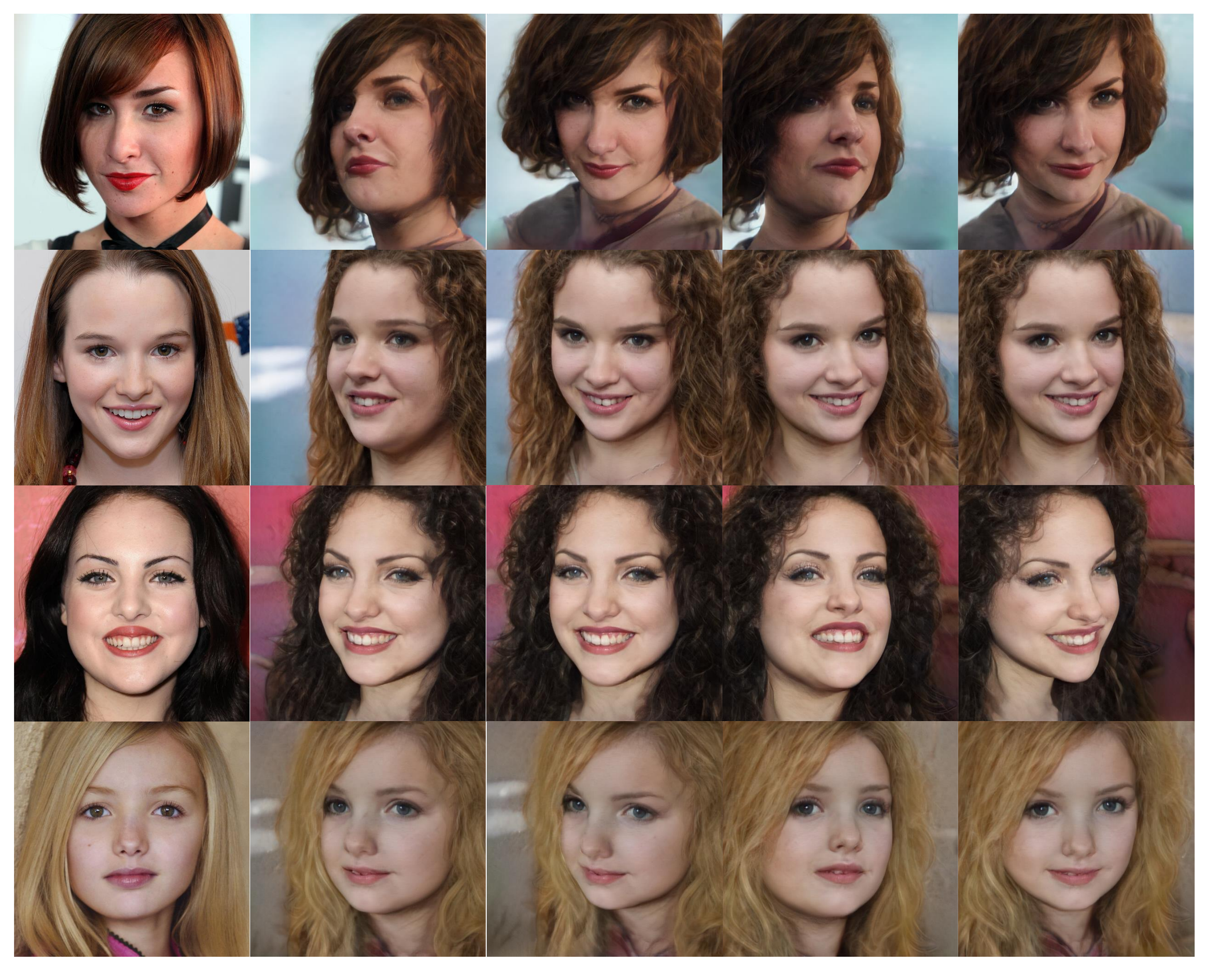}
   \caption{The wavy hair editing results obtained by PREIM3D.
   The first column is the input image.
   }
   \label{fig:vr_wavyhair}
\end{figure*}

We provide a large number of inversion and editing results produced by PREIM3D in Figure~\ref{fig:vr_inversion} to ~\ref{fig:vr_wavyhair}

\end{document}